\documentclass[twocolumn,letterpaper]{IEEEAerospaceCLS}

\usepackage[]{graphicx}  
\usepackage[colorlinks=true, urlcolor=blue, linkcolor=black, citecolor=black]{hyperref} 

\usepackage{booktabs} 
\usepackage{array} 
\usepackage{amssymb}
\usepackage{amsmath}
\usepackage{epsfig}
\usepackage{epstopdf}
\usepackage{subfigure}
\usepackage{subcaption}
\usepackage{booktabs}
\usepackage[table]{xcolor}
\usepackage{lipsum}

\usepackage{cite}
\usepackage{graphics, framed}
\usepackage{url}
\usepackage{multimedia}
\usepackage{paralist}
\usepackage[printonlyused]{acronym}
\usepackage{units}
\usepackage{balance}
\usepackage{multirow}
\usepackage{color,soul}
\usepackage[ruled,vlined]{algorithm2e}
\usepackage{dsfont}

\newcommand{\FGR}[1]{Fig.~\ref{#1}}

\newcommand{\SEC}[1]{Section~\ref{#1}}
\newcommand{\APP}[1]{Appendix~\ref{#1}}
\newcommand{\TAB}[1]{Table~\ref{#1}}
\newcommand{\ALG}[1]{Algorithm~\ref{#1}}

\newcommand{\ignore}[1]{}  
\usepackage{longtable,array} 
\newcolumntype{C}[1]{>{\centering\arraybackslash}p{#1}}

\begin{document}
\title{PIKAN: Physics-Inspired Kolmogorov–Arnold Networks for Explainable UAV Channel Modelling}

\author{%
Kürşat Tekbıyık\\ 
Poly-Grames Research Centre\\Polytechnique Montr\'eal, GERAD \& Mila\\
kursat.tekbiyik@polymtl.ca
\and 
Güneş Karabulut Kurt\\
Poly-Grames Research Centre\\Polytechnique Montr\'eal\\
gunes.kurt@polymtl.ca
\and 
Antoine~Lesage-Landry\\
Poly-Grames Research Centre\\ Polytechnique Montr\'eal, GERAD \& Mila\\
antoine.lesage-landry@polymtl.ca
\thanks{\footnotesize 979-8-3315-7360-7/26/$\$31.00$ \copyright2026 IEEE}              
}

\maketitle

\thispagestyle{plain}
\pagestyle{plain}

\hyphenation{pi-pe-li-nes}

\maketitle

\thispagestyle{plain}
\pagestyle{plain}

\begin{abstract}
Unmanned aerial vehicle (UAV) communications demand accurate yet interpretable air-to-ground (A2G) channel models that can adapt to non-stationary propagation environments. While deterministic models offer interpretability and deep learning (DL) models provide accuracy, both approaches suffer from either rigidity or a lack of explainability. To bridge this gap, we propose the Physics-Inspired Kolmogorov–Arnold Network (PIKAN) that embeds physical principles (e.g., free-space path loss, two-ray reflections) into the learning process. Unlike physics-informed neural networks (PINNs), PIKAN is more flexible for applying physical information because it introduces them as adaptable inductive biases. Thus, it enables a more flexible training process. Experiments on UAV A2G measurement data show that PIKAN achieves comparable accuracy to DL models while providing symbolic and explainable expressions aligned with propagation laws. Remarkably, PIKAN achieves this performance with only 232 parameters, making it up to 37 times lighter than multilayer perceptron (MLP) baselines with thousands of parameters, without sacrificing correlation with measurements and also providing symbolic expressions. These results highlight PIKAN as an efficient, interpretable, and scalable solution for UAV channel modelling in beyond-5G and 6G networks.

\end{abstract} 

\vspace{-0.2in}
\tableofcontents

\section{Introduction}\label{sec:intro}

Unmanned aerial vehicles (UAVs) have recently become indispensable components of future wireless networks due to their rapid deployment, three-dimensional mobility, and potential to extend coverage in both urban and rural environments~\cite{zeng2019accessing}. As UAVs can serve as aerial base stations, relays, or mobile users, enabling applications ranging from disaster recovery and environmental monitoring to logistics, surveillance, and broadband access in underserved areas~\cite{li2018uav}, the integration of UAVs into 6G and beyond networks is therefore expected to play a key role in ensuring ubiquitous and resilient connectivity.

A critical factor for realizing the potential of UAV communications is the development of accurate air-to-ground (A2G) channel models. Traditional terrestrial channel models, such as those defined in 3GPP~TR~38.901~\cite{3gpp38901}, are tailored for urban macrocell, microcell, or rural terrestrial scenarios and fail to capture the unique propagation conditions of UAV links, including altitude-dependent fading, high line-of-sight (LoS) probability, strong Doppler shifts, and rapid non-stationarity~\cite{khawaja2019survey}. This motivates the design of new UAV-specific channel models that balance physical interpretability and prediction accuracy.

\subsection{Related Works}

UAV channel modelling has been studied using deterministic, statistical, and empirical approaches. Empirical measurement campaigns have provided valuable insights into large-scale path loss and fading, with comparative studies recommending suitable models depending on frequency, altitude, and environment~\cite{al2014modeling, matolak2016air, xiao2025measurements}. More recently, cluster-based modelling and machine learning methods have been employed to characterize time-varying propagation structures, enabling improved accuracy in 6G UAV-to-ground channels~\cite{zhang2024ml}. Moreover, \cite{khawaja2019survey}~highlights the limitations of reusing terrestrial models and emphasizes the need for UAV-specific approaches.

In parallel, deep learning (DL) approaches have emerged as alternatives for channel modelling. Architectures such as generative adversarial networks (GANs) have been proposed to generate realistic wireless channels and enrich training datasets~\cite{xiao2022channelgan}. Beyond generative models, physics-informed neural networks (PINNs)~\cite{raissi2019pinn} have been introduced to embed physical equations directly into the training process. PINNs promote consistency with known physical models; however, they can be overly rigid, struggling to adapt when empirical data deviates from theoretical assumptions. Other DL-based methods~\cite{seah2024empirical, chen2022deep, lee2019path} have demonstrated strong performance but remain black-boxes in nature, offering no or very limited interpretability for underlying propagation mechanisms.

These developments highlight an important gap: existing DL-based channel modelling either lacks interpretability or imposes rigid physical constraints, which limit learning flexibility when the data deviates from the theory or known phenomena. This motivates the search for an approach that achieves both high accuracy and symbolic interpretability while maintaining adaptability to real-world UAV channels.

\subsection{Motivations}

While deterministic and empirical models provide physical interpretability, they often lack flexibility and fail to generalize across diverse UAV scenarios. Conversely, DL-based approaches achieve high prediction accuracy but sacrifice transparency, making it difficult to explain or validate learned channel behaviours. This trade-off is particularly problematic for UAV communications, where explainable models are essential for ensuring reliability, safety, and their adoption by standardization bodies.

To address this issue, recent developments such as Kolmogorov--Arnold Networks (KANs)~\cite{liu2024} offer a promising direction. KANs retain the expressive aspect of neural networks while producing symbolic, interpretable representations of learned functions. Leveraging this capability allows us to combine the performance of data-driven models with the explainability from physical principles.

Unlike traditional symbolic regression methods, KANs offer a continuous, gradient-based search in the function space. KANs therefore provide interpretable intermediate representations that can be inspected and refined during training. Symbolic regressions are typically fragile, returning only a success or failure without revealing interpretable intermediate steps; whereas KANs retain their performance even when the underlying target function is not perfectly symbolic. By combining splines and neural networks, they offer high accuracy, better scalability, and improved interpretability. This makes KANs a more flexible and transparent option than traditional symbolic regression approaches.

\subsection{Contributions}

In this paper, we introduce the \textit{Physics-Inspired Kolmogorov–Arnold Network} (PIKAN), a novel KAN-based framework for modelling UAV channels. The main contributions are summarized as follows:

\begin{enumerate}[{C}1.]
    \item We propose a channel modelling approach that integrates the interpretability of KANs with the guidance from physical laws, e.g., two-ray reflections, path-loss exponents, etc.
    \item We provide a path loss model considering other environmental parameters along with symbolic expressions because employing DL enables to learn the relation between path loss and other parameters than the distance and carrier frequency, which are not considered in most of the deterministic models.
    \item Unlike PINNs~\cite{raissi2019pinn}, where physical phenomena appear directly in the loss function, our approach employs no such regularizer. Instead, we configure the KAN by fixing its activation functions to symbolic relations from physical expressions, such as the free space path loss (FSPL). Thus, we refer to the model as \textit{inspired} to distinguish it from PINNs.
    
    \item We demonstrate through experiments on the UAV channel dataset~\cite{10757825} that PIKAN achieves competitive accuracy compared to black-box DL models while providing symbolic, explainable expressions that align with physical intuition as well. This makes PIKAN a novel middle ground between black-box learning and physics-based modelling.
\end{enumerate}

In summary, UAV communication requires new perspectives in channel modelling that requires both accuracy and interpretability. By combining KANs with physics-inspired training, our PIKAN framework provides a combined solution with improved performance and symbolic expression of UAV channels.

Next, we introduce KANs in~\SEC{sec:kan}. In~\SEC{sec:dataset}, we explain the dataset used in this study. \SEC{sec:training} addresses the training approaches and introduces the model structure. In~\SEC{sec:experiments}, we evaluate the trained models and introduce the PIKAN approach along with symbolic models. Also, we provide benchmarking results comparing with both deterministic and ML-based models. \SEC{sec:conclusion} concludes the paper, discussing future works. In the Appendices, we provide additional information regarding the model, dataset, and results.

\section{Kolmogorov--Arnold Networks}\label{sec:kan}
KANs are a neural network architecture inspired by the Kolmogorov--Arnold representation theorem rather than the classical universal approximation theorem that motivates multilayer perceptrons (MLPs)~\cite{liu2024, poeta2024benchmarking}. While MLPs rely on fixed nonlinear activation functions applied to nodes with linear weights on edges, KANs reverse this perspective: nodes only perform summation, whereas edges carry learnable univariate nonlinear functions. This structural change allows KANs to achieve better interpretability and favourable approximation properties compared to MLPs~\cite{ji2024comprehensive}.

The theoretical foundation of KANs is provided by the Kolmogorov--Arnold representation theorem~\cite{liu2024}. It states that any continuous multivariate function can be expressed as a finite composition of continuous univariate functions and addition. Let $f:[0,1]^n \to \mathbb{R}$ be a continuous function. Then, there exist continuous univariate functions $\varphi_{q,p}: [0,1]\to\mathbb{R}$ and $\Phi_q:\mathbb{R}\to\mathbb{R}$ such that
\begin{equation}
    f(x_1,\dots,x_n) = \sum_{q=1}^{2n+1} \Phi_q \!\left( \sum_{p=1}^n \varphi_{q,p}(x_p) \right).
    \label{eq:KA_theorem}
\end{equation}

Basically, the only true multivariate operation is the outer summation; all other functions can be constructed from univariate transformations and summations. This representation motivates the design of KANs.

Given input--output pairs $\{(\mathbf{x}_i, y_i)\}_{i=1}^{I}$, where $I \in \mathbb{N}$ is the size of the dataset and $\mathbf{x}_i \in \mathbb{R}^{P}$ with $P \in \mathbb{N}$ being the number of input features, a KAN models the mapping $\text{KAN}(\mathbf{x}) \approx f(x) = y$ by parameterizing the univariate functions in \eqref{eq:KA_theorem} as learnable splines~\cite{liu2024}. A KAN is structured in layers, with each layer consisting of a matrix of univariate functions:

\begin{equation}
    \mathbf{\Phi}^l = \{\varphi^l_{j,i}\}, \quad 
    i=1,\dots,n_l,\ j=1,\dots,n_{l+1},
    \label{eq:KAN_layer}
\end{equation}
where $n_l \in \mathbb{N}$ denotes the number of nodes in the $l$-th layer. The forward propagation rule is
\begin{equation}
    x^{l+1}_j = \sum_{i=1}^{n_l} \varphi^l_{j,i}(x^l_i), 
    \quad j=1,\dots,n_{l+1}.
    \label{eq:KAN_forward}
\end{equation}

In matrix form, a KAN layer can be expressed as
\begin{equation}
    \mathbf{x}^{l+1} = \mathbf{\Phi}^l(x^l),
\end{equation}
where $\mathbf{\Phi}^l \in \mathbb{R}^{n_l \times n_{l+1}}$ is a matrix of learnable univariate spline functions. Stacking $L$ such layers yields
\begin{equation}
    \text{KAN}(\mathbf{x}) = (\mathbf{\Phi}^{L-1} \circ \mathbf{\Phi}^{L-2} \circ \cdots \circ \mathbf{\Phi}^0)(\mathbf{x}).
    \label{eq:KAN_full}
\end{equation}

For comparison, an MLP is defined as
\begin{equation}
    \text{MLP}(\mathbf{x}) = (\mathbf{W}^{L-1}\circ \sigma \circ \mathbf{W}^{L-2}\circ \cdots \circ \sigma \circ \mathbf{W}^0)(\mathbf{x}),
\end{equation}
where $\mathbf{W}^l \in \mathbb{R}^{n_l \times n_{l+1}}$ are linear weight matrices and $\sigma$ are fixed nonlinear activation functions. KANs differ fundamentally by replacing linear weights with nonlinear learnable functions.

The approximation ability of KANs can be quantified as follows. Let suppose each $\varphi^l_{j,i}$ is parameterized by B-splines of order $k \in \mathbb{N}$ on a grid of size $G \in \mathbb{N}$. Then, one can show that the approximation error decays as
\begin{equation}
    \| f - \text{KAN}(\mathbf{x}) \|_{C^m} \leq C G^{-(k+1-m)}, \quad 0 \leq m \leq k,
    \label{eq:KAN_error_bound}
\end{equation}
for some constant $C > 0$ depending on $f$, where $C^m$ denotes the norm measuring the magnitude of derivatives up to order $m$~\cite{liu2024}. This indicates that KANs can achieve scaling exponents $\alpha = k+1$ in neural scalability, significantly outperforming MLPs that often saturate at lower exponents~\cite{somvanshi2024survey, poeta2024benchmarking}.

\section{The Dataset}\label{sec:dataset}       
The dataset used in this study is collected with a custom-built software-defined radio (SDR) channel sounder integrated into the Aerial Experimentation and Research Platform for Advanced Wireless (AERPAW) testbed~\cite{10757825, raouf2025wireless}. The measurements take place in a rural environment with minimal interference and few obstacles, which allows isolation of the key A2G channel characteristics.

A total of nine UAV flights are carried out using a hexacopter equipped with a portable receiver node that includes a USRP B210, a GNSS-disciplined oscillator (GNSS-DO) board, and an Intel NUC for control and data collection. A fixed transmitter node, mounted on a 10 m tower and equipped with a USRP B210 and a GNSS-DO board, provides the reference signal. Both ends operate at a 56 MHz sampling rate, which is the maximum bandwidth supported by the USRP B210.

The transmitted waveform is a Zadoff–Chu (ZC) sequence with a length of 401 symbols and root index 200, repeated four times per frame. Measurements are taken at 4 Hz over three carrier frequencies within the Citizens Broadband Radio Service (CBRS) band (3564, 3620, and 3686 MHz) and three constant flight altitudes at 30, 60, and 90 m. The UAV maintains a flight speed of 5 m/s along a straight 500 m path while keeping its yaw orientation fixed. The effective transmit power is 19 dBm.

The dataset provides key wireless channel metrics, including path loss, channel impulse response (CIR), received signal power, and delay spreads. The controlled rural setting and systematic variations in altitude and frequency create a comprehensive basis for developing and evaluating A2G channel models under line-of-sight (LoS)-dominant conditions. The full dataset and sounder code are openly available in~\cite{usrpchannelsounder2024}.

In this study, we only consider the path loss measurements to create a path loss model based on KAN. Details and figures regarding the dataset can be found in~\APP{sec:dataset_appendix}. We split the dataset into training and test subsets based on predefined time windows. Specifically, samples whose time indices fall within the intervals [500,800], [1100,1400], and [1800,2100] are selected for the test set, while the remaining samples constitute the training set. This ensures that the test data originates from distinct temporal segments of the measurement campaign as seen in~\FGR{fig:aerpaw_dataset}.

After the split, we apply robust scaling~\cite{amorim2022scaling} to both subsets as
\begin{equation}
\bar{x}_{i,j} =
\frac{x_{i,j} - \operatorname{med}\!\bigl(\mathbf{X}_{:,j}\bigr)}
     {\operatorname{IQR}\!\bigl(\mathbf{X}_{:,j}\bigr)},
\end{equation}
where \(\bar{(\cdot)}\) is the scaled feature, \(\mathbf{X}\) is the training-data matrix,
\(\mathbf{X}_{:,j}\) denotes its \(j\)-th feature column,
\(\operatorname{med}(\cdot)\) is the median,  
and \(\operatorname{IQR}(\cdot)\) is the interquartile range (third minus first quartile)
computed from the training data only. Using a robust scaler when standardizing the features avoids from the impacts of outliers. The scaling parameters are computed exclusively from the training data to avoid information leakage. These parameters are then used to transform both the training and test datasets. The scaler parameters are given in~\TAB{tab:scaler_stats}. 

\begin{table}[!t]
\centering
\caption{Robust scaling statistics.}
\label{tab:scaler_stats}
\renewcommand{\arraystretch}{1.15}
\begin{tabular}{l r r l}
\hline\hline
\textbf{Feature} & \textbf{Median} & \textbf{IQR (Q3--Q1)} & \textbf{Unit} \\
\hline
$f_{\mathrm{c}}$                  &     3620 & 122       & MHz \\
$d_{\mathrm{hor}}$                &   44.321 & 215.340   & m \\
$d_{\mathrm{ver}}$               &   18.052 &  35.888   & m \\
$\alpha_{\mathrm{AoA}}$  &    0.399 &   0.948   & rad \\
$\beta_{\mathrm{AoA}}$   &   -1.599 &   1.083   & rad \\
$\alpha_{\mathrm{AoD}}$  &   -0.399 &   0.948   & rad \\
$\beta_{\mathrm{AoD}}$   &    1.543 &   1.083   & rad \\
\hline\hline
\end{tabular}%
\end{table}

\section{Training Parameters and Approach}\label{sec:training}
\begin{figure*}[!t]
    \centering
    \subfigure[]{
        \label{fig:kan_training_all}
        \includegraphics[width=0.45\linewidth, page=1]{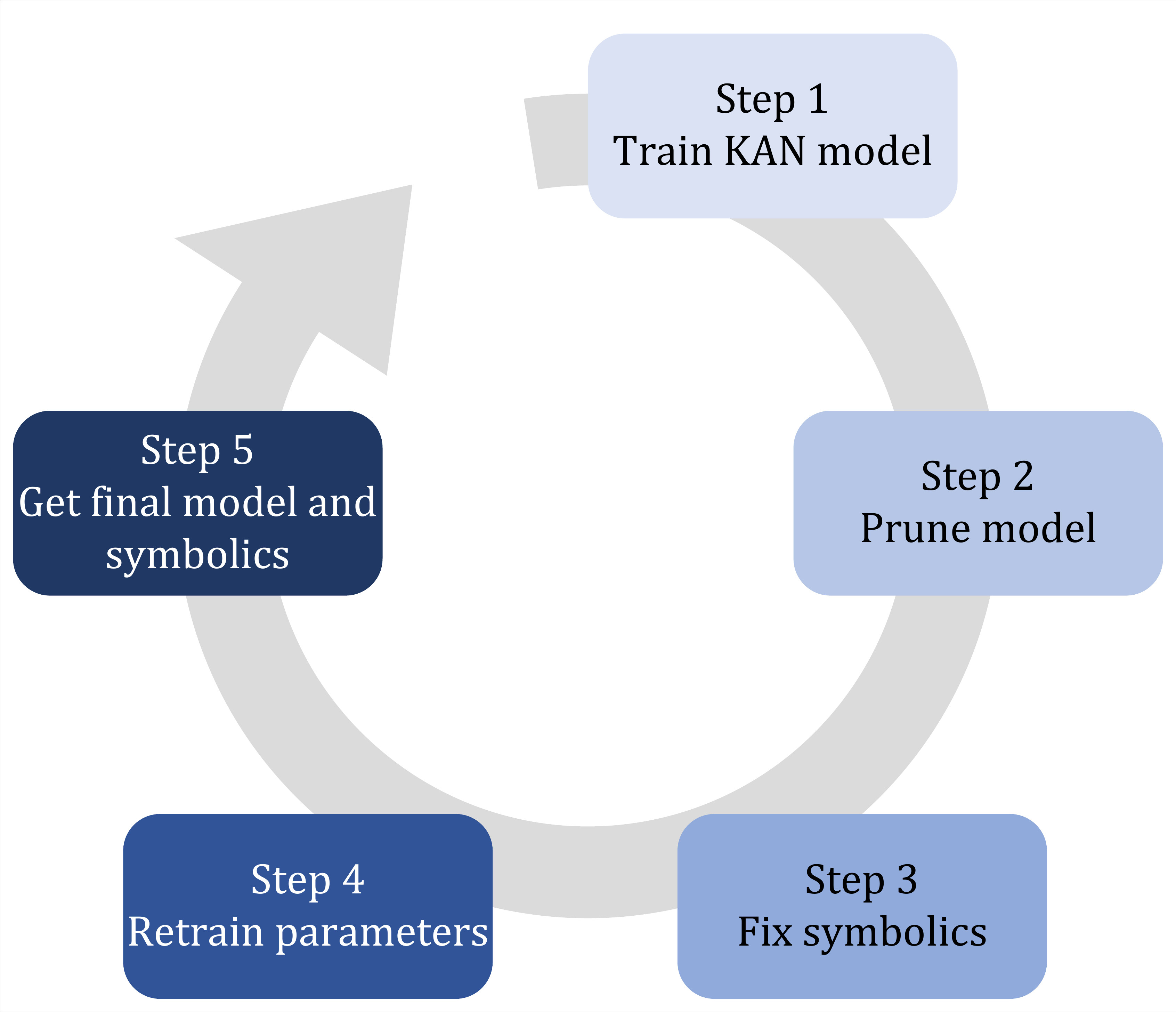}}
    \quad
    \subfigure[]{
        \label{fig:kan_training_pi}
        \includegraphics[width=0.45\linewidth, page=2]{./figs/utils/kan_training_v2}}
    \caption{Training pipelines for Kolmogorov–Arnold Networks: 
            (a) baseline pruning and symbolic refinement; 
            (b) extended pipeline with \textit{physics-inspired} symbolic fix.}
    \label{fig:kan_training}
\end{figure*}
We define the KAN with an architecture of [$N$,~1,~1], which means that the number of hidden units equals the number of input features, $N$, and there is a single scalar output. The output represents the measured path loss expressed in decibels (dB). This compact design highlights the effect of KAN-specific basis-function tuning rather than network depth. It should be noted that, unlike symbolic regression, KANs perform a continuous gradient-based search and reveal interpretable intermediate functions, and accuracy even when there is no exact symbolic expression~\cite{liu2024}. 

To evaluate prediction performance, we train and test the model while varying both the input-feature set and the key hyperparameters grid size, $G$ and~$k$. In KANs, width corresponds to the number of basis functions that define each activation; although our hidden layer contained only one unit, the effective width is determined by these basis functions, with larger widths increasing representational capacity but also the training cost and the risk of overfitting. The grid specifies the knot points for the B-splines that shape each activation function, so a finer grid can capture more detailed structure in the data, but at the expense of higher computation and potential for overfitting. The $k$ parameter sets the degree of the B-splines and thus the smoothness of the activations, where higher values yield smoother, more stable functions but may reduce the flexibility of learning outliers where the practical data does not obey theoretical or known physical phenomena. All other training details followed our standard setup: we used the Adam optimizer with a learning rate of $1\times10^{-4}$, a mini-batch size of $64$, $200$ training steps, a loss function of root mean square error (RMSE), and a fixed random seed of $123$ for reproducibility. 

We consider two complementary training pipelines for symbolic function acquisition, as illustrated in~\FGR{fig:kan_training}. In the first approach shown in~\FGR{fig:kan_training_all}, the model is initially trained (Step~1) and subsequently pruned to reduce complexity (Step~2). Symbolic functions are then adjusted (Step~3) to enhance interpretability, after which the network parameters are retrained (Step~4). The process concludes with the extraction of the final symbolic model and parameters (Step~5).

In the second approach in~\FGR{fig:kan_training_pi}, we extend the pipeline by introducing a physics-inspired symbolic correction step (Step~3), followed by parameter retraining (Step~4). Additional symbolic refinements are performed (Step~5) before retraining once more (Step~6). Finally, the best combination of parameters and symbolic functions is obtained (Step~7). This iterative process allows us to integrate physical priors into the symbolic representation while maintaining model accuracy.

\begin{algorithm}[t]
\SetArgSty{textnormal}
\SetInd{0.5em}{1em} 
\caption{Gradient Vanishing Resilient Adaptive Learning Rate Backoff}
\label{alg:safe_train}
\KwIn{Model $M$, dataset $\mathcal{D}$, total steps $S$, segment length $s$, 
initial learning rate $\eta_0$, minimum learning rate $\eta_{\min}$, 
backoff factor $\beta \in (0,1)$}
\KwOut{Best model state, validation loss}

Initialize $\eta \leftarrow \eta_0$, $\text{Loss}_{\text{best}} \leftarrow \infty$ \;
Save initial model state $M_{\text{safe}}$ \;

\While{steps $<$ $S$}{
    Train $M$ on $\mathcal{D}$ for $s$ steps with lr = $\eta$\;
    Compute test loss $\text{Loss}_{\text{test}}$\;
    
    \eIf{$\text{Loss}_{\text{test}} < \text{Loss}_{\text{best}}$}{
        $\text{Loss}_{\text{best}} \leftarrow \text{Loss}_{\text{test}}$\;
        Save $M_{\text{safe}} \leftarrow M$\;
    }{
        Restore $M \leftarrow M_{\text{safe}}$\;
        $\eta \leftarrow \beta \eta$\;
        \If{$\eta < \eta_{\min}$}{\textbf{break}}
    }
}
\Return{$M_{\text{safe}}, \text{Loss}_{\text{best}}$}
\end{algorithm}

During retraining (Steps 4–5 in~\FGR{fig:kan_training}), one of the major challenges is gradient vanishing, particularly after the pruning and symbolic correction. To mitigate this, we employ a segmented training routine with adaptive learning-rate backoff. The procedure is summarized in~\ALG{alg:safe_train}, where training is performed in segments, training loss is continuously monitored, and the learning rate is adaptively reduced upon detecting divergence. This ensures numerical stability and convergence to a solution. We set the total steps, segment length, initial learning rate, and backoff factor as $200$, $25$, $3 \times 10^{-4}$, and $0.3$,~respectively.~These~approaches~are~used~in~\SEC{sec:experiments}.

\section{Experiments}\label{sec:experiments}
As noted in~\SEC{sec:training}, we perform a grid search on hyperparameters and input features such as carrier frequency, distance, and so on to identify the best model–data combination. A full detailed discussion on the available features is provided in~\APP{sec:dataset_appendix}. Sweeping $k~\in~\{2,3,4\}$ and grid size $G~\in~\{5,10,15,20\}$, the lowest test RMSE is achieved at \textbf{$k=3$} and \textbf{$G=5$}. The most influential input features are the center frequency $f_{\mathrm{c}}$, horizontal distance $d_{\mathrm{hor}}$, vertical distance $d_{\mathrm{ver}}$, and angles of arrival $\alpha_{\mathrm{AoA}}$ and $\beta_{\mathrm{AoA}}$. Complete grid-search results are given in~\TAB{tab:model_inputs_sweep}. As shown in \FGR{fig:grid_search}, as the grid $G$, increases, the training performance generally improves while the test performance decreases. This is because the higher $G$ leads to a finer fit on the train data. Increasing $k$ behaves as a regularization because it leads to smoother splines. Our model selection is based on the balance between train performance and avoiding overfitting. 

\begin{figure*}[!t]
    \centering
    \subfigure[]{
        \label{figs:train_last_mse}
        \includegraphics[width=0.45\linewidth]{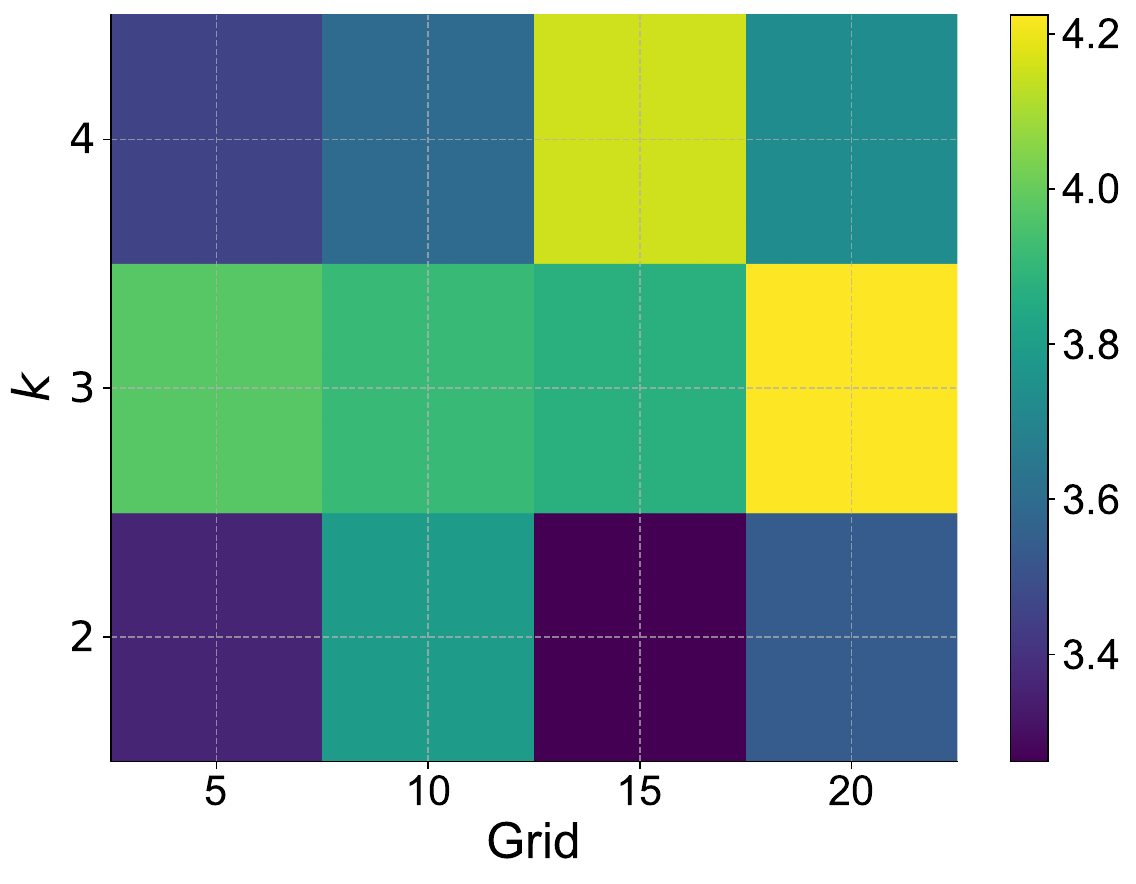}}
    \quad
    \subfigure[]{
        \label{fig:val_last_mse}
        \includegraphics[width=0.45\linewidth]{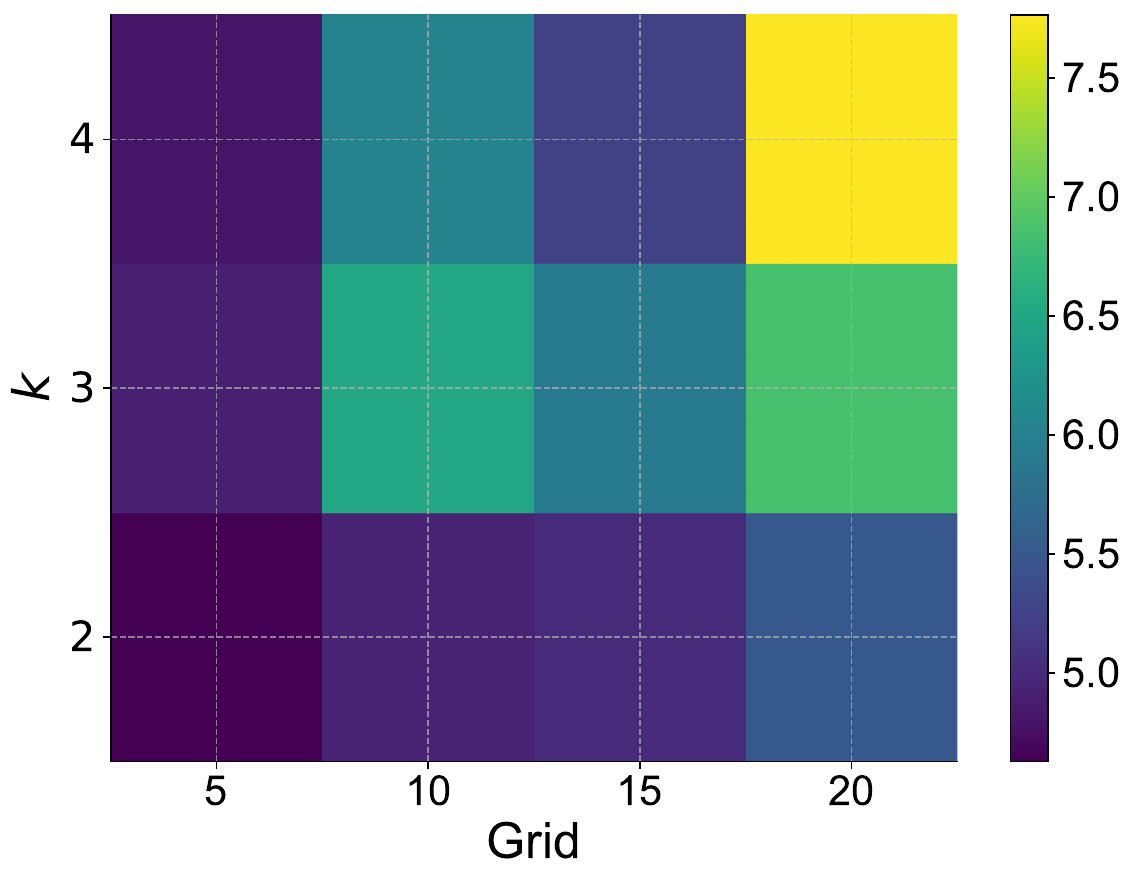}}
        \quad
    \subfigure[]{
        \label{fig:val_mae}
        \includegraphics[width=0.45\linewidth]{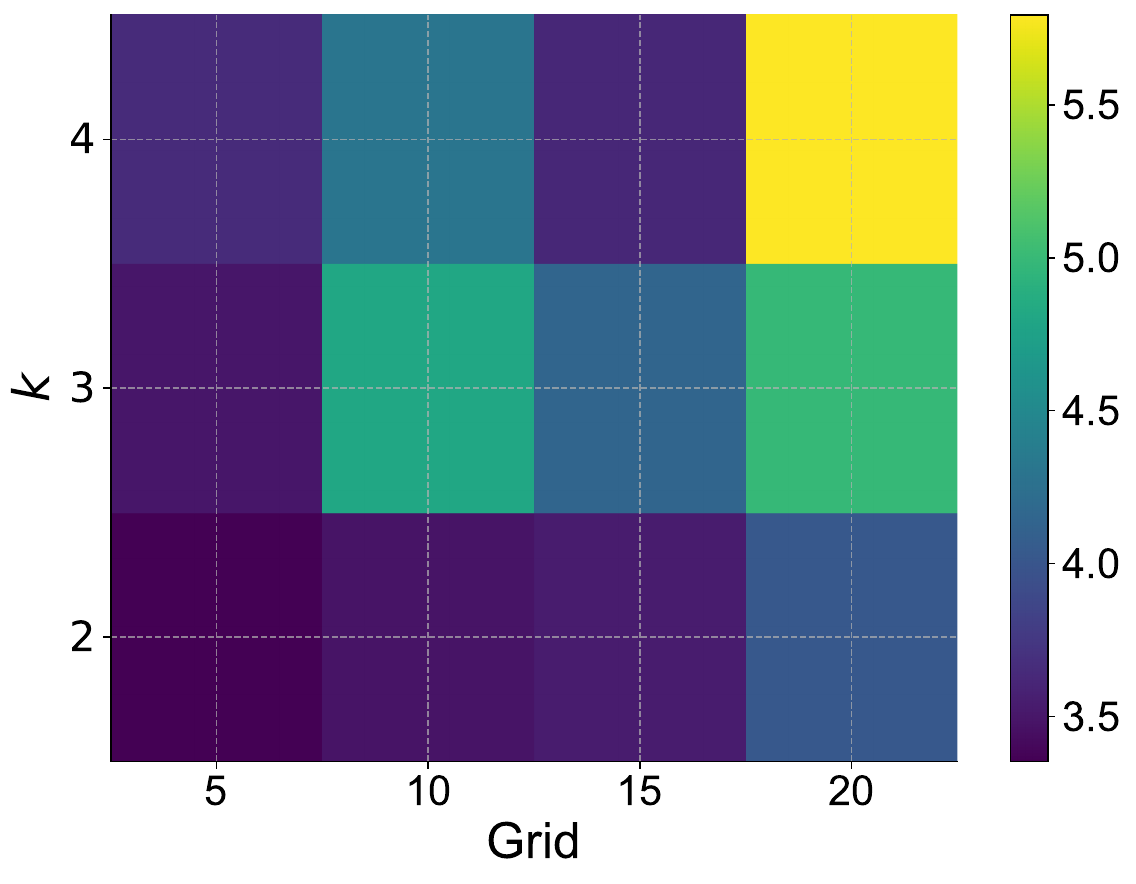}}
    \quad
    \subfigure[]{
        \label{fig:val_corr}
        \includegraphics[width=0.45\linewidth]{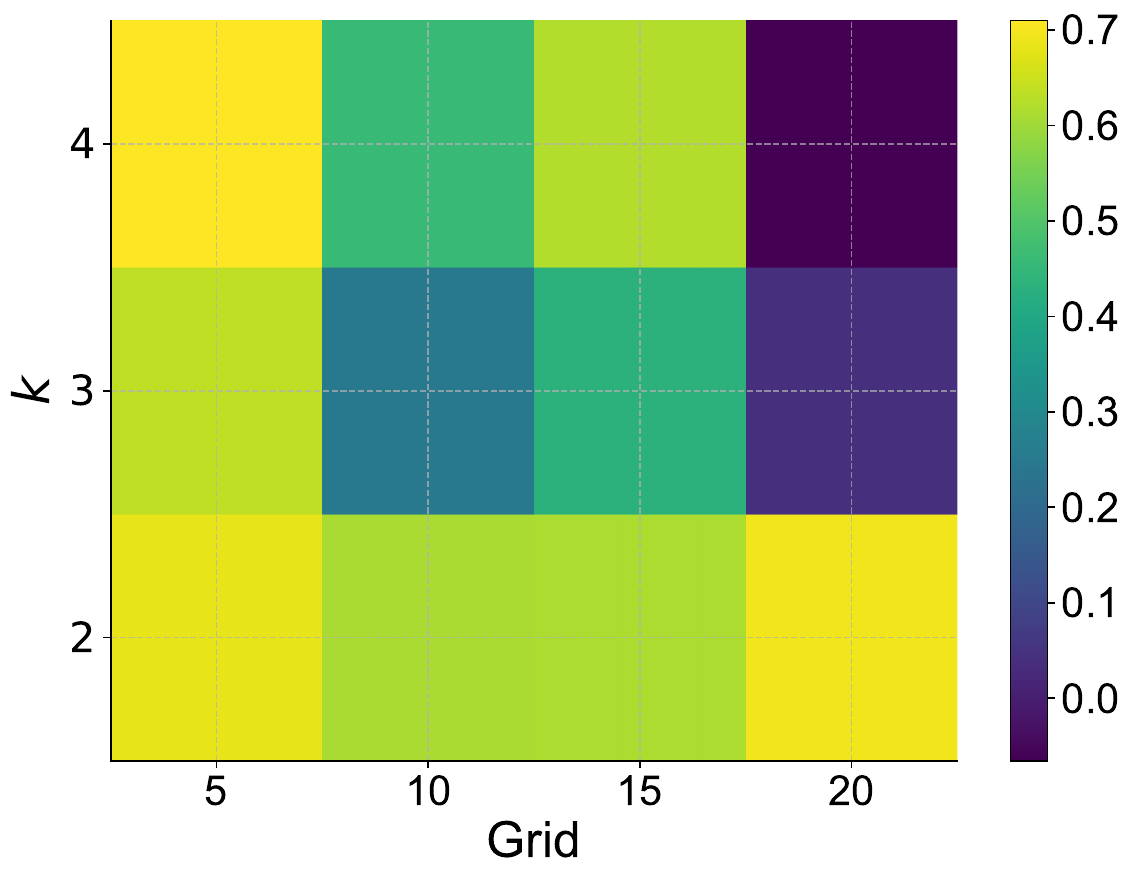}}
    \caption{Grid search performance results: (a) training RMSE, (b) test RMSE, (c) test MAE, (d) test correlation scores.}
    \label{fig:grid_search}
\end{figure*}

\subsection{From Model to Symbolic Function}\label{sec:model2sym}
As described in~\SEC{sec:training} and illustrated in~\FGR{fig:kan_training}, we first prune unnecessary edges and nodes of the base model shown in~\FGR{fig:base_model} to reduce the network’s complexity and eliminate unimportant features, thereby improving interpretability before attempting symbolification. The performance of each stage is summarized in~\TAB{tab:kan_eval}, where pruning yields a comparable performance to the base model while simplifying the structure.

After the pruning, we employ the symbolification interface~\cite{liu2024}, which allows fixing activations to candidate symbolic functions. Specifically, we define a symbolic library as
\begin{equation}
    \left\{ (\cdot), (\cdot)^2, (\cdot)^3, (\cdot)^4, \exp, \log, \sqrt{\cdot}, \tanh, \sin, \tan, \lvert \cdot \rvert \right\},  
    \label{eq:lib}
\end{equation}
where no restrictions are imposed on the selection. The best-fitting symbolic form is automatically chosen by iteratively matching the pre-activations $\mathbf{x}$ and post-activations $y$ with affine-transformed symbolic functions $cf(a\mathbf{x}+b)+d$, using grid search over $(a,b,c,d) \in \mathbb{R}$ and linear regression. This procedure enables the network to yield compact and interpretable symbolic expressions while preserving predictive accuracy.

The resulting symbolic function corresponding to the pruned base model is expressed as:
\begin{equation}
\begin{aligned}
L &= -5.25\, \overline{\alpha}_{\mathrm{AoA}}
   + 1.85\, \overline{\beta}_{\mathrm{AoA}}
   + 2.17\, \overline{\beta}_{\mathrm{AoD}} \\
  &\quad + 8.69\, \left(\overline{f}_{\mathrm{c}} + 0.12\right)^{2} \\
  &\quad - 2.09\, \sin\left(1.72\, \overline{d}_{\mathrm{hor}} + 1.23\right) \\
  &\quad - 1.64\, \sin\left(1.39\, \overline{d}_{\mathrm{ver}} + 1.99\right) \\
  &\quad - 7.23\, \sin\left(5.62\, \overline{\alpha}_{\mathrm{AoD}} - 8.00\right) + 73.53,
\end{aligned}
\end{equation}
where $\overline{(\cdot)}$ denotes the scaled parameter as explained in~\SEC{sec:dataset} and given in~\TAB{tab:scaler_stats}.

To further improve the symbolic model, we retrain the symbolic function to update its affine parameters using the safe training algorithm \ALG{alg:safe_train}. The retrained expression is given as:
\begin{equation}
\begin{aligned}
L &= -5.23\, \overline{\alpha}_{\mathrm{AoA}}
   + 1.39\, \overline{\beta}_{\mathrm{AoA}}
   + 2.16\, \overline{\beta}_{\mathrm{AoD}} \\
  &\quad + 8.47\, \left(\overline{f}_{\mathrm{c}} + 0.12\right)^{2} \\
  &\quad - 2.11\, \sin\left(1.73\, \overline{d}_{\mathrm{hor}} + 1.22\right) \\
  &\quad - 1.62\, \sin\left(1.39\, \overline{d}_{\mathrm{ver}} + 1.98\right) \\
  &\quad - 7.20\, \sin\left(5.60\, \overline{\alpha}_{\mathrm{AoD}} - 8.01\right) + 73.78.
\end{aligned}
\end{equation}

It is observed in~\TAB{tab:kan_eval} that while the symbolic model achieves interpretability, its performance slightly degrades compared to the base and pruned models. Retraining recovers part of this loss but does not fully match the prediction performance of the non-symbolic KAN.

\begin{figure}[!t]
    \centering
    \includegraphics[width=0.7\linewidth]{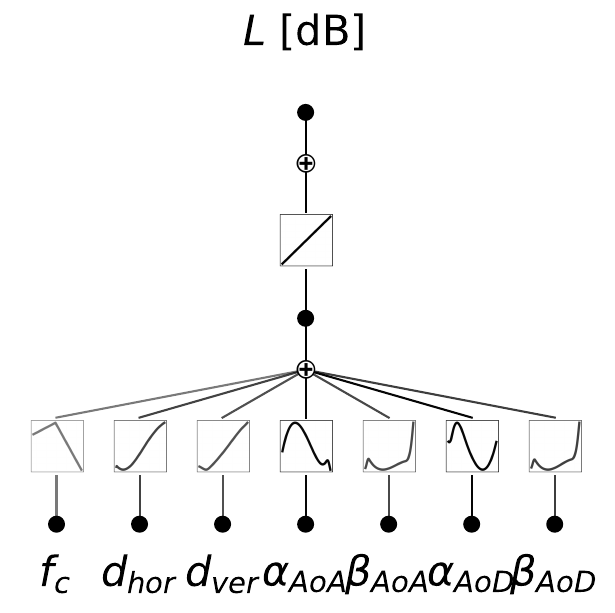}
    \caption{The base KAN structure before symbolification. The output node corresponds to the predicted path loss $L$ [dB], while the input nodes represent the selected significant features: center frequency $f_{\mathrm{c}}$, horizontal distance $d_{\mathrm{hor}}$, vertical distance $d_{\mathrm{ver}}$, and angles of arrival/departure $(\alpha_{\mathrm{AoA}}, \beta_{\mathrm{AoA}}, \alpha_{\mathrm{AoD}}, \beta_{\mathrm{AoD}})$.}
    \label{fig:base_model}
\end{figure}

\begin{table*}[!t]
\centering
\caption{Key evaluation metrics for KAN models (the best models highlighted with shaded rows).}
\label{tab:kan_eval}
\begin{tabular}{lcccc}
\hline\hline
\textbf{KAN Models} & \textbf{MAE} & \textbf{RMSE} & $\mathbf{R^2}$ & $\mathbf{\rho}$ \\
\hline
Base KAN                              & 3.083 & 3.984 & 0.557  & 0.746 \\
Symbolic Base KAN                          & 4.069 & 4.898 & 0.330  & 0.600 \\
Retrained Symbolic KAN               & 4.058 & 4.883 & 0.334  & 0.596 \\
PIKAN-FSPL                       & 3.070 & 3.966 & \textbf{0.561}  & 0.749 \\
\rowcolor{gray!20}
Retrained PIKAN-FSPL             & \textbf{2.993} & 4.053 & 0.541  & \textbf{0.750} \\
Symbolic PIKAN-FSPL              & 4.020 & 4.973 & 0.310  & 0.590 \\
Retrained Symbolic PIKAN-FSPL    & 3.751 & 4.759 & 0.368  & 0.677 \\
\rowcolor{orange!20}
PIKAN-2R                    & 3.068 & \textbf{3.963} & \textbf{0.561} & \textbf{0.750} \\
Symbolic PIKAN-2R         & 4.120 & 4.949 & 0.316  & 0.594 \\
Retrained Symbolic PIKAN-2R         & 8.816 & 9.545 & -1.544 & 0.736 \\
\hline\hline
\end{tabular}%
\end{table*}

\subsection{Physics-Inspired KANs}\label{sec:pin}

In the previous section, we process the base KAN model step by step to obtain symbolic activation functions, resulting in an interpretable model. This section is devoted to drawing inspiration from physical models and applying it to reform the KAN's activation functions. Unlike physics-informed networks, where the physical phenomenon directly appears in the loss function, our approach does not use such a regularizer. Instead, we configure the KAN by fixing its activation functions to symbolic relations from known physical expressions like the two-ray model. Hence, we use the term \textit{inspired}, as the model to differentiate it from PINNs.

After the pruning, we fix certain activation functions to predetermined symbolic functions, considering the FSPL model as given below:
\begin{equation}
    L_{\mathrm{FSPL}} = 20 \log_{10}\left(\frac{4 \pi d f_{\mathrm{c}}}{c}\right),
\end{equation}
where $d = \sqrt{d_{\mathrm{hor}}^2 + d_{\mathrm{ver}}^2}$ and $c$ is the speed of light. So, we fix activation functions to symbolic functions as:
\begin{equation}
(i,j,k) \mapsto
\begin{cases}
(0,1,0) \mapsto {(\cdot)}^{2}, \\
(0,2,0) \mapsto {(\cdot)}^{2}, \\
(1,0,0) \mapsto \log(\cdot),
\end{cases}
\end{equation}

\begin{figure}[!t]
    \centering
    \includegraphics[width=0.7\linewidth]{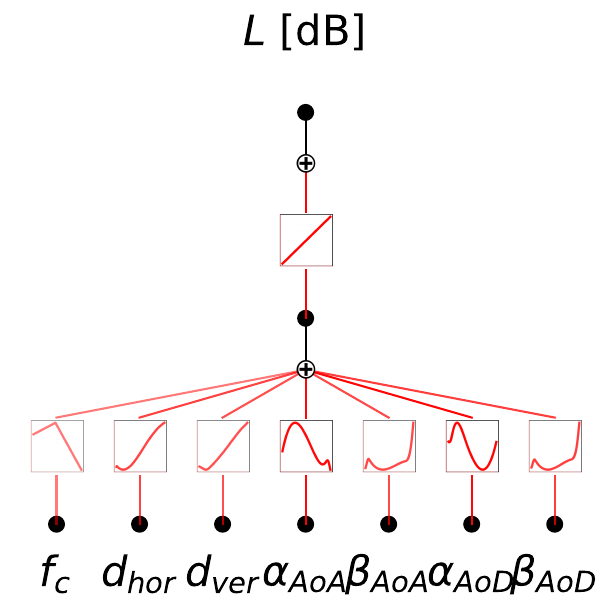}
    \caption{The pruned KAN structure after symbolification. Each node is associated with the best-fitting symbolic activation function chosen from the predefined library. \textcolor{red}{Red} color denotes the symbolic activation functions.}
    \label{fig:model_symbolic}
\end{figure}

where $(i,j,k)$ denotes the layer index, input index, and output index in the KAN model, and $\mapsto$~is symbolic fixing operator. PIKAN model is illustrated in~\FGR{fig:model_pruned_pin}. This inspiration reduces mean absolute error (MAE) from $3.083$ to $3.07$ and provides insight about the model. We remark that this model is not fully symbolic yet. Similarly to the previous section, for PIKAN, we retrain the affine parameters and obtain a retrained PIKAN model with the lowest MAE of $2.993$ in the test data.

When fixing the remaining activation functions to symbolic functions from the library given in~\eqref{eq:lib}, we get the fully symbolic and explainable PIKAN model described in~\eqref{eq:symbolic_PIKAN_fspl}, where \textcolor{purple}{purple} coloured terms are physics-inspired symbolic fixing. Retraining its affine parameters reduces the MAE from $4.02$ to $3.75$. The resulting path loss estimator is provided in~\eqref{eq:symbolic_PIKAN_fspl_retrained}.

\begin{figure}[!t]
    \centering
    \includegraphics[width=0.7\linewidth]{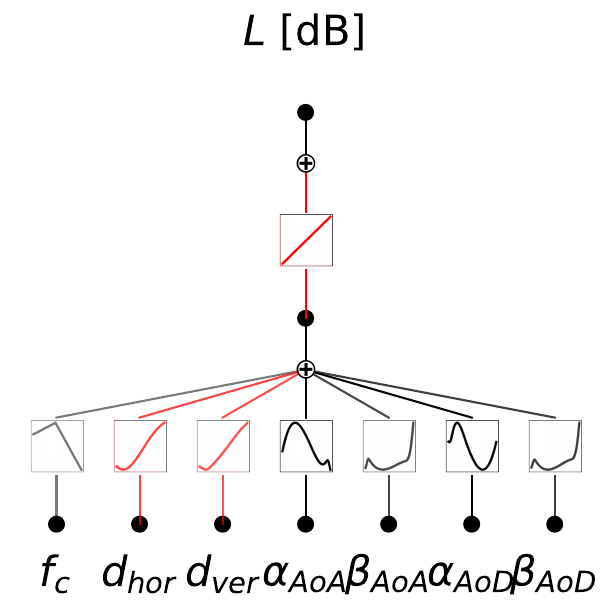}
    \caption{The PIKAN structure inspired by free space path loss. Only functions related to distances and outer function are fixed. \textcolor{red}{Red} color denotes the symbolic activation functions.}
    \label{fig:model_pruned_pin}
\end{figure}

\begin{figure*}[t]
\centering
\fbox{%
\parbox{0.97\textwidth}{%
\begin{equation}
\begin{aligned}
L &= 3.69 \times 10^{3}\, \textcolor{purple}{\log}\Big(
   -1.60 \times 10^{-2}\, \overline{\alpha}_{\mathrm{AoA}}
   + 4.83 \times 10^{-3}\, \overline{\beta}_{\mathrm{AoA}}
   + 5.72 \times 10^{-3}\, \overline{\beta}_{\mathrm{AoD}} \\
  &\quad + 2.39 \times 10^{-2}\, \left(\overline{f}_{\mathrm{c}} + 1.17 \times 10^{-1}\right)^{2} 
  -2.98 \times 10^{-3}\, \left(\overline{d}_{\mathrm{hor}} + 1.63 \times 10^{-1}\right)^{\textcolor{purple}{2}} \\
  &\quad - 4.20 \times 10^{-3}\, \left(3.99 \times 10^{-1}\, \overline{d}_{\mathrm{ver}} + 1\right)^{\textcolor{purple}{2}}
  - 1.96 \times 10^{-2}\, \sin\left(5.62\, \overline{\alpha}_{\mathrm{AoD}} + 4.58\right)
   + 10.2 \Big) \\
  &\quad - 8.51 \times 10^{3}
\end{aligned}
\label{eq:symbolic_PIKAN_fspl}
\end{equation}
}%
}
\end{figure*}

\begin{figure*}[t]
\centering
\fbox{%
\parbox{0.97\textwidth}{%
\begin{equation}
\begin{aligned}
L &= 3.69 \times 10^{3}\, \textcolor{purple}{\log}\Big(
   -1.57 \times 10^{-2}\, \overline{\alpha}_{\mathrm{AoA}}
   + 4.51 \times 10^{-3}\, \overline{\beta}_{\mathrm{AoA}}
   + 5.54 \times 10^{-3}\, \overline{\beta}_{\mathrm{AoD}} \\
  &\quad + 2.37 \times 10^{-2}\, \left(\overline{f}_{\mathrm{c}} + 1.17 \times 10^{-1}\right)^{2} 
  - 6.39 \times 10^{-3}\, \left(\overline{d}_{\mathrm{hor}} + 1.63 \times 10^{-1}\right)^{\textcolor{purple}{2}} \\
  &\quad - 3.04 \times 10^{-3}\, \left(3.97 \times 10^{-1}\, \overline{d}_{\mathrm{ver}} + 1\right)^{\textcolor{purple}{2}}
  - 1.92 \times 10^{-2}\, \sin\left(5.60\, \overline{\alpha}_{\mathrm{AoD}} + 4.56\right)
   + 10.2 \Big) \\
  &\quad - 8.51 \times 10^{3}
\end{aligned}
\label{eq:symbolic_PIKAN_fspl_retrained}
\end{equation}
}%
}
\end{figure*}

Besides FSPL, we utilize two-ray model~\cite{haykin2005}:
\begin{equation}
L_{\mathrm{2Ray}} = 40 \log_{10}(d) - 10 \log_{10}\!\left(G_{\mathrm{t}} G_{\mathrm{r}} h_{\mathrm{t}}^{2} h_{\mathrm{r}}^{2}\right),
\end{equation}
to reshape the base KAN model and obtain the PIKAN-2R model, where $G_{\mathrm{t}}$ and $G_{\mathrm{r}}$ antenna gains, and $h_{\mathrm{t}}$ and $h_{\mathrm{r}}$ denote the altitude of UAV and height of ground terminal. We observe that path loss is inversely proportional to the square of the horizontal distance.  Thus, we can fix the activation functions to symbolics as follows: 

\begin{equation}
(i,j,k) \mapsto
\begin{cases}
(0,1,0) \mapsto {(\cdot)}^{4}, \\
(0,2,0) \mapsto {(\cdot)}^{-2}, \\
(1,0,0) \mapsto \log(\cdot).
\end{cases}
\end{equation}

This insight reduces the MAE from $3.083$ to $3.068$, which is lower than what both the base and the PIKAN-FSPL models achieve. Also, it provides the lowest RMSE and the best correlation with the measurements. Similar to PIKAN-FSPL, fixing the remaining activation functions to symbolics in the library creates a fully symbolic PIKAN-2R model as described in~\eqref{eq:symbolic_PIKAN_tworay} with an MAE of $4.12$. However, retraining the affine parameters leads to a higher error, which remains an open issue for our next studies.

\begin{figure*}[t]
\centering
\fbox{%
\parbox{0.97\textwidth}{%
\begin{equation}
\begin{aligned}
L &= 3.69 \times 10^{3}\, \textcolor{purple}{\log}\Big(
   -1.45 \times 10^{-2}\, \overline{\alpha}_{\mathrm{AoA}}
   + 5.10 \times 10^{-3}\, \overline{\beta}_{\mathrm{AoA}}
   + 6 \times 10^{-3}\, \overline{\beta}_{\mathrm{AoD}} \\
  &\quad - 2.40 \times 10^{-2}\, \left(\overline{f}_{\mathrm{c}} + 1.17 \times 10^{-1}\right)^{2}
  + 1.39 \times 10^{-3}\, \left(4.09 \times 10^{-1}\, \overline{d}_{\mathrm{hor}} + 1\right)^{\textcolor{purple}{4}} \\
  &\quad - 2.00 \times 10^{-2}\, \sin\!\left(5.62\, \overline{\alpha}_{\mathrm{AoD}} - 8.00\right)
  + 1.02 \times 10^{1}
   + \dfrac{2.23 \times 10^{-2}}{\left(1 - 8.55 \times 10^{-2}\, \overline{d}_{\mathrm{ver}}\right)^{\textcolor{purple}{2}}}
   \Big) \\
  &\quad - 8.51 \times 10^{3}
\end{aligned}
\label{eq:symbolic_PIKAN_tworay}
\end{equation}
}%
}
\end{figure*}

\subsection{Benchmarking}\label{sec:benchmarking}

In this subsection, we benchmark the proposed KAN- and PIKAN-based models against conventional deterministic models and baseline MLP architectures. The objective is to assess prediction accuracy, interpretability, and model complexity.

\subsubsection{Deterministic Models}
We first evaluate four deterministic models widely used in the literature: Urban Macro (UMa) LoS, UMa Non-Line-of-Sight (NLoS), ITU-S, and flat Earth two-ray (FE2R).  

\begin{figure*}[t]
\centering
\begin{minipage}{0.95\textwidth}
\begin{equation}
\label{eq:uma_los}
L_{\mathrm{UMa,LoS}} =
\begin{cases}
28 + 22 \log_{10}(d) + 20 \log_{10}(f_{\mathrm{c}}) + 4, & 10 < d \leq d_{\mathrm{BP}}, \\[6pt]
28 + 40 \log_{10}(d) + 20 \log_{10}(f_{\mathrm{c}}) - 9 \log_{10}\!\big(d_{\mathrm{BP}}^{2} + (d_{\mathrm{ver}} - 10)^{2}\big) + 4, & d_{BP} < d \leq 5000
\end{cases}
\end{equation}

\begin{equation}
\label{eq:uma_nlos}
L_{\mathrm{UMa,NLoS}} =
\max\!\Big(L_{\mathrm{UMa,LoS}},\; 8.44 + 39.08 \log_{10}(d) + 20 \log_{10}(f_{\mathrm{c}})\Big) + 6,\;  10 < d \leq 5000.
\end{equation}
\end{minipage}
\end{figure*}

The UMa LoS and NLoS models~\cite{3gpp38901} are defined in~\eqref{eq:uma_los} and ~\eqref{eq:uma_nlos}, respectively, where $d_{\mathrm{BP}}$ is the breakpoint distance expressed as:
\begin{equation}
    d_{\mathrm{BP}} = 18 \pi (d_{\mathrm{ver}} - 1) f_{\mathrm{c}}. \label{eq:uma_bp}
\end{equation}

Another widely used path loss model is the ITU-S model~\cite{itu1411} given by:
\begin{equation}
    L_{\mathrm{ITU-S}} = A f^{B} d_\mathrm{v}^{D} (\Theta + E)^{G}, 
    \label{eq:itu_s}
\end{equation}
where $d_\mathrm{v}$ is the vegetation depth in meters, $\Theta$ is the elevation angle in degrees, and $A, B, D, E, \text{and}, G$ are model parameters. We used the parameters given in~\cite{xiao2025measurements}.

The last model is the FE2R model~\cite{parsons2000} expressed as:
\begin{equation}
    L_{\mathrm{FE2R}} = -20 \log_{10} \!\left( \frac{\lambda}{4 \pi d}
        \left| \frac{1}{d} + \frac{R {\mathrm{e}}^{-j \frac{2\pi(d' - d)}{\lambda}}}{d'} \right| \right), 
    \label{eq:fe2r}
\end{equation}
with
\begin{equation}
    R = \frac{\frac{d_{\mathrm{ver}}}{d} - \sqrt{\xi_{\mathrm{r}}- \frac{\left(\frac{d_{\mathrm{hor}}}{d}\right)^2}{\xi_{r}}}}{\frac{d_{\mathrm{ver}}}{d} + \sqrt{\xi_{r} - \frac{\left(\frac{d_{\mathrm{hor}}}{d}\right)^2}{\xi_{r}}}},  
\end{equation}
and 
\begin{equation}
    d' = \sqrt{d_{\mathrm{hor}}^{2} + (d_{\mathrm{ver}} + 2h_{\mathrm{r}})^{2}}, 
\end{equation}
where $d'$ is the reflected path distance, $\lambda$ is the wavelength, $h_{\mathrm{r}}$ is the receiver antenna height, and $\xi_{\mathrm{r}}$ is the relative permittivity. We use $h_{\mathrm{r}} = 10$ as given in~\cite{10757825} and $\xi_{\mathrm{r}} = 15$ as in~\cite{xiao2025measurements}.

\TAB{tab:det_eval} presents the evaluation of deterministic baselines alongside the proposed symbolic KAN and PIKAN models. Classical deterministic models such as UMa LoS and UMa NLoS show limited accuracy, with large error magnitudes and correlation coefficients that are either low ($\rho = 0.398$ for UMa LoS) or unstable despite reduced RMSE. Other empirical models, such as FE2R and ITU-S, provide moderate improvements, but their errors remain an order of magnitude larger than those of the proposed methods.

By contrast, the symbolic KAN model significantly improves predictive accuracy, achieving an MAE of $4.058$ and RMSE of $4.883$, which are markedly lower than all traditional deterministic baselines. This improvement demonstrates the effectiveness of a symbolic approximation within the framework.

Building further on this approach, the symbolic PIKAN model with FSPL guidance provides the best overall performance, reducing the error to $3.751$ while also reaching the highest correlation coefficient $\rho = 0.677$. This improvement underscores the benefit of incorporating physics priors into the model structure, enabling a better balance between physical interpretability and statistical accuracy.

\begin{figure}[!t]
    \centering
    \includegraphics[width=\linewidth]{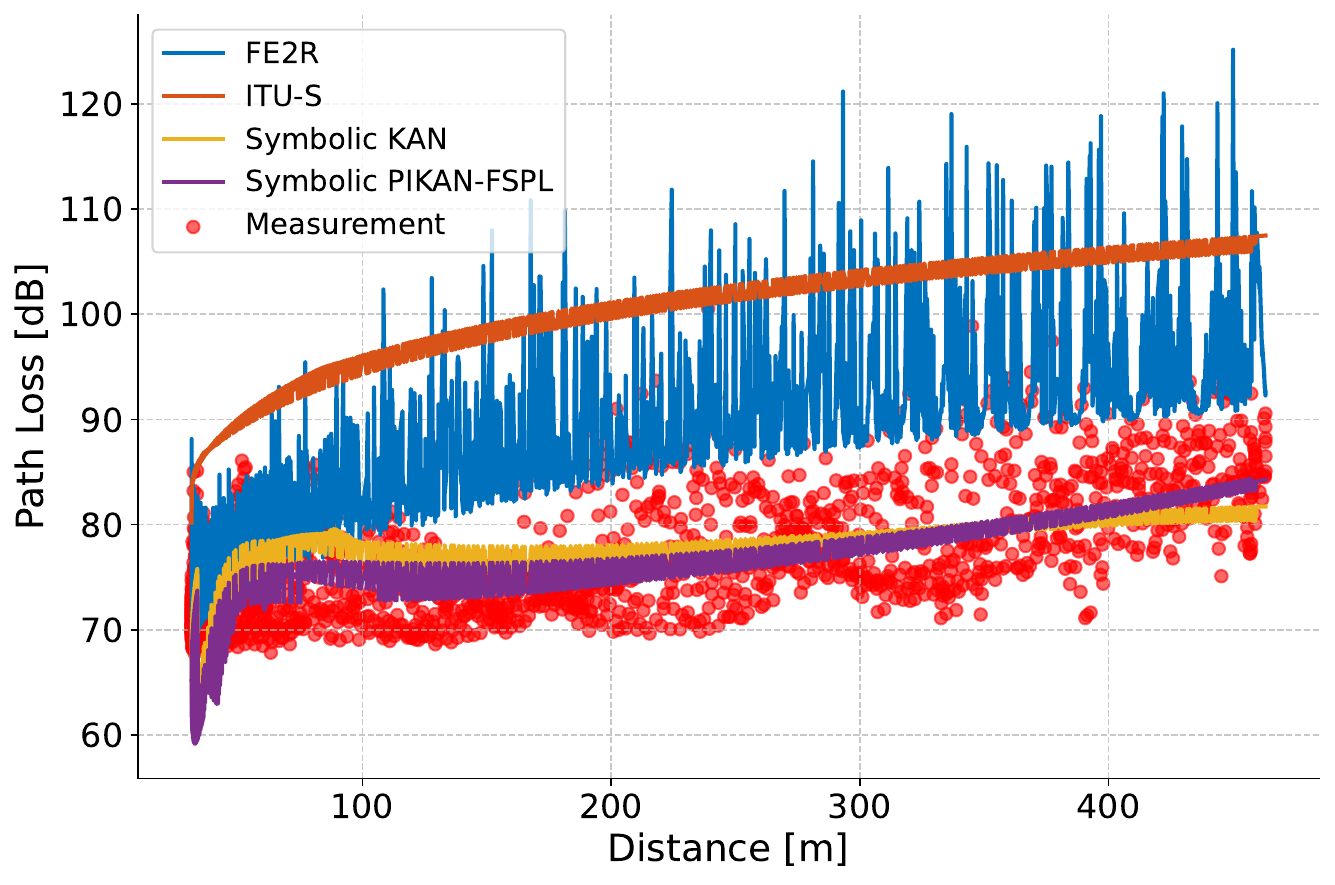}
    \caption{Comparison of deterministic models with the proposed symbolic KAN and PIKAN frameworks. The PIKAN–FSPL model achieves the best overall accuracy across all metrics, highlighting the benefits of embedding physics-inspired structures.}
    \label{fig:pl_vs_models}
\end{figure}

Although the symbolic PIKAN–2R model shows slightly weaker performance than its FSPL counterpart, it still maintains a significant acccuracy over traditional baselines, confirming the high performance of the proposed PIKAN methodology. Overall, these results highlight the superiority of the proposed symbolic KAN and PIKAN approaches over conventional deterministic models in capturing A2G path loss behaviour with both higher accuracy and improved generalization. Moreover, the improved accuracy of the proposed symbolic KAN and PIKAN–FSPL models over conventional deterministic baselines is further illustrated in~\FGR{fig:pl_vs_models}, where they more closely follow the measurement data compared to FE2R and ITU-S.

These results validate the contribution of physics-inspired symbolic modelling in achieving interpretable yet accurate channel representations. It therefore supports the main motivation of this work.

\begin{table}[!t]
\centering
\caption{Key evaluation metrics for deterministic models and symbolic versions of KAN and PIKAN models (best values highlighted with shaded rows).}
\label{tab:det_eval}
\resizebox{\linewidth}{!}{%
\begin{tabular}{lccc}
\hline\hline
\textbf{Models} & \textbf{MAE} & \textbf{RMSE} & $\mathbf{\rho}$ \\
\hline
UMa LoS   & 63.831  & 68.719  & 0.398 \\
UMa NLoS  & 90.194  & 91.162  & 0.675 \\
FE2R      & 9.474 & 12.047 & 0.573 \\
ITU-S     & 19.405  & 20.377 & 0.661 \\
Symbolic Base KAN            & 4.058 & 4.883  & 0.596 \\
\rowcolor{gray!20}
Symbolic PIKAN-FSPL    & \textbf{3.751} & \textbf{4.759}  & \textbf{0.677} \\
Symbolic PIKAN-2R & 4.120 & 4.949  & 0.594 \\
\hline\hline
\end{tabular}
}%
\end{table}

\subsubsection{Neural Network Baselines} \label{sec:nn_benchmark}

As a learning-based reference, several MLP models of varying depth and width are trained. Please refer to~\APP{sec:mlp_training} for details regarding the training procedure. \TAB{tab:mlp_eval} summarizes the evaluation metrics for baseline MLPs and the proposed KAN and PIKAN approaches. Among the MLP configurations, deeper or wider architectures generally reduce the MAE and RMSE while slightly improving the correlation coefficient $\rho$. For example, the 64$\times$64$\times$32$\times$16 architecture achieves the best performance among MLPs, reaching an MAE of $2.736$, RMSE of $3.895$, and $\rho = 0.767$. However, this comes at the cost of a parameter count of $8529$, which is nearly 37$\times$ larger than that of the proposed KAN models.

The proposed KAN and PIKAN models offer a more balanced tradeoff between accuracy and complexity. The base KAN already outperforms most MLPs, achieving an MAE of $3.083$, RMSE of $3.984$, and $\rho~=~0.746$ with only $232$ parameters. This represents more than two orders of magnitude fewer parameters compared to the best-performing MLP while maintaining competitive accuracy. Incorporating physical priors through the PIKAN framework further enhances performance. Both FSPL- and two-ray-based PIKAN models achieve the lowest errors among the proposed methods (MAE $\approx 3.07$, RMSE $\approx 3.96$) and the highest correlation ($\rho = 0.750$), without increasing the parameter count.

These results corroborate that embedding physics-inspired structures into the KAN framework yields consistently better generalization compared to purely data-driven MLPs. Moreover, the serious reduction in parameters highlights the suitability of KAN and PIKAN models for resource-constrained implementations, such as UAV channel modelling tasks, where both interpretability and efficiency are critical.

\begin{table}[!t]
\centering
\caption{Key evaluation metrics and parameter counts for MLP and KAN-based models (best values highlighted in bold).}
\label{tab:mlp_eval}
\resizebox{\linewidth}{!}{%
\begin{tabular}{lccc}
\hline\hline
\textbf{Models} & \textbf{MAE} & \textbf{RMSE} & \textbf{\# Params} \\
\hline
MLP $32\times32$        & 3.421 & 4.501  & 1441 \\
MLP $128\times64$       & 3.210 & 4.331 & 10049 \\
MLP $256\times128\times64$           & 2.986 & 4.237  & 52353 \\
MLP $64\times64\times32\times16$           & \textbf{2.736} & \textbf{3.895}  & 8529 \\
Base KAN           & 3.083 & 3.984  & \textbf{232} \\
PIKAN-FSPL        & 3.070 & 3.966  & \textbf{232} \\
PIKAN-2R     & 3.068 & 3.963  & \textbf{232} \\

\hline\hline
\end{tabular}%
}
\end{table}

\section{Conclusions and Future Work}\label{sec:conclusion}
This work introduces physics-inspired Kolmogorov-Arnold network (PIKAN) for unmanned aerial vehicle (UAV) channel modelling, which combines the symbolic interpretability of KANs with the inductive guidance of physics-based models. Benchmarking against both deterministic baselines and deep multilayer perceptrons (MLPs) shows that PIKAN provides the best balance between accuracy, complexity, and interpretability. For example, it achieves mean absolute error (MAE) $\approx$ 3.07 and correlation, $\rho \approx 0.75$ with 37$\times$ fewer parameters than the best-performing MLP. By providing symbolic path loss expressions that conform to physical intuition, PIKAN provides a new middle ground between physics-based models and black-box deep learning (DL) models.

For future works, we are extending PIKANs in several directions:
\begin{itemize}
    \item \textit{Retraining strategies:} We are investigating stable retraining of the PIKAN-2R model to enhance its predictive performance without overfitting.
    \item \textit{Deeper architectures:} We are developing multi-layer PIKAN structures to flexibly integrate more complex propagation effects. 
    \item \textit{Cross-dataset experiments:} Despite the scarcity of open UAV datasets, we are conducting cross-dataset experiments to validate our approach on a larger scale.
    \item \textit{Benchmarking with PINN:} Besides MLP, we are planning to benchmark the proposed approach with PINN to show the potential learning flexibility. 
    \item \textit{Foundation models:} We are working toward a PIKAN-based foundation model for wireless channels with aims for providing  a scalable and adaptable representation across frequency bands, environments, and mobility conditions.
\end{itemize}

In summary, by combining physical inspiration with symbolic interpretability and lightweight design, PIKANs pave the way for efficient and explainable channel models that could accelerate UAV and 6G standardization.

\acknowledgements
The authors would like to thank Prof.~İsmail~Güvenç and Dr.~Anıl~Gürses for providing access to the UAV A2G measurement dataset used in this study.

This work is supported by the Natural Sciences and Engineering Research Council (NSERC) of Canada Alliance grant ALLRP~579869-22 ("Artificial Intelligence Enabled Harmonious Wireless Coexistence for 3D Networks (3D- HARMONY)").

\bibliographystyle{IEEEtran}
\bibliography{main}

\thebiography
\begin{biographywithpic}
{K{\"{u}}r{\c{s}}at Tekb{\i}y{\i}k}{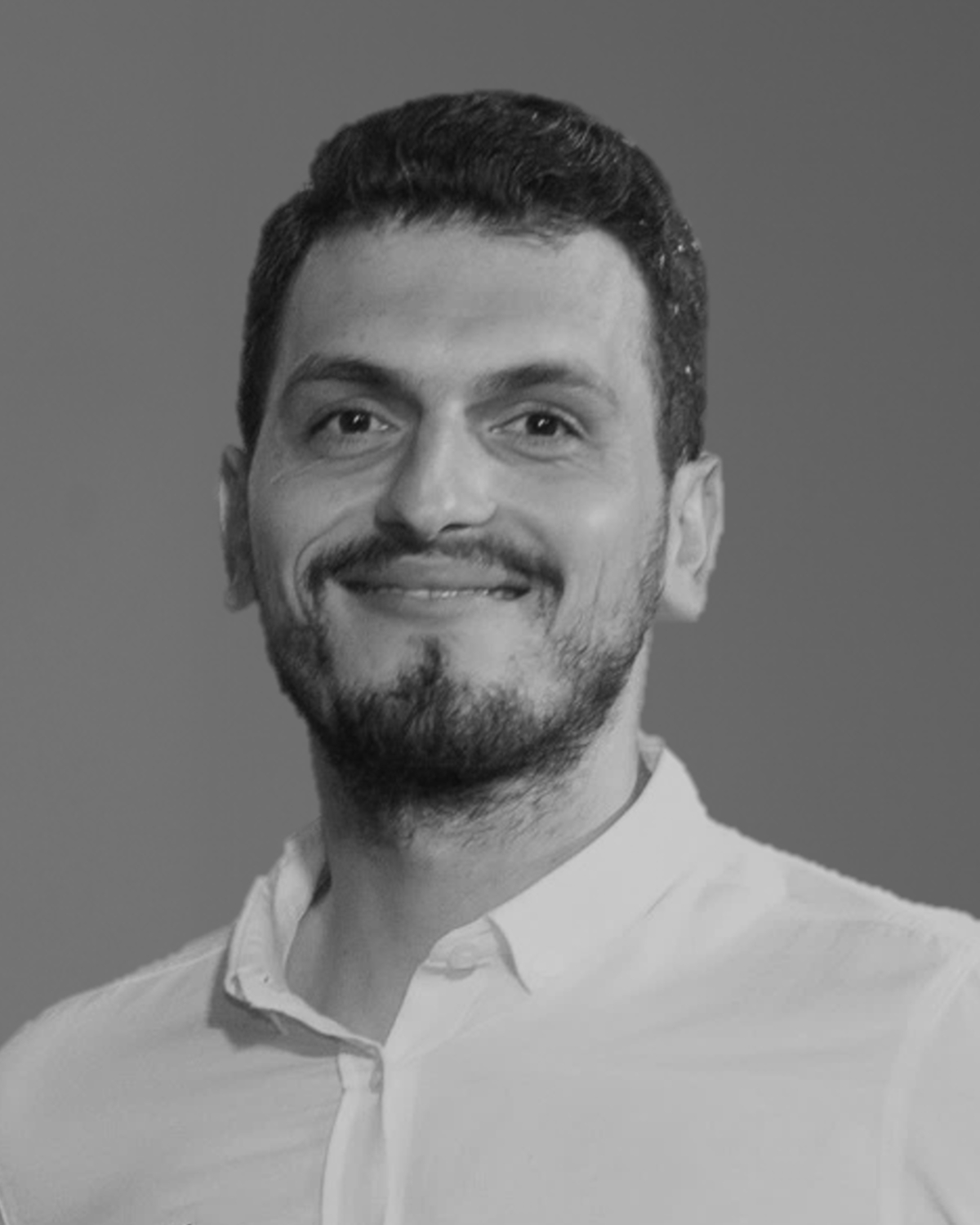} (Member,~IEEE) received his B.Sc., M.Sc., and Ph.D. degrees in Telecommunications Engineering from Istanbul Technical University, Turkey, in 2017, 2019, and 2024, respectively. His previous works cover areas ranging from signal intelligence systems to terahertz communications. His research interests include nonterrestrial networks and machine learning applications in wireless communications.
\end{biographywithpic} 

\begin{biographywithpic}
{Güneş Karabulut Kurt}{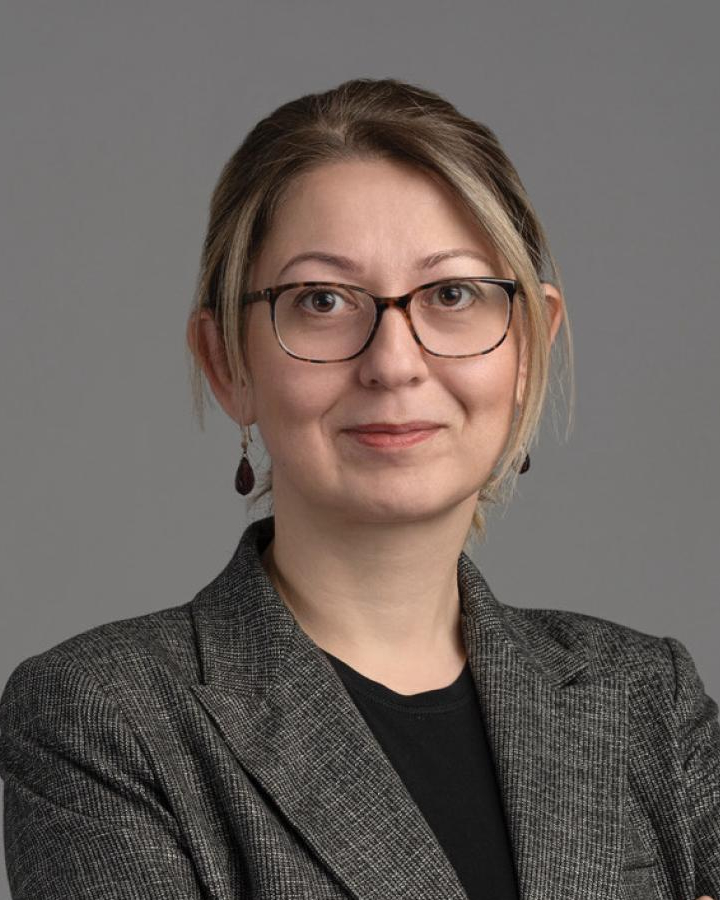}
(Senior~Member,~IEEE) is a Canada Research Chair (Tier 1) in New Frontiers in Space Communications and Full Professor at Polytechnique Montréal, Montréal, QC, Canada. She is also an adjunct research professor at Carleton University. Gunes received the B.S. degree with high honors in electronics and electrical engineering from Bogazici University, Istanbul, Turkiye, in 2000 and the M.A.Sc. and the Ph.D. degrees in electrical engineering from the University of Ottawa, ON, Canada, in 2002 and 2006, respectively. She worked in different technology companies in Canada and Turkiye between 2005 and 2010. From 2010 to 2021, she was a professor at Istanbul Technical University. Gunes is a Marie Curie Fellow and has received the Turkish Academy of Sciences Outstanding Young Scientist (TÜBA-GEBIP) Award in 2019. She is serving as the secretary of the IEEE Satellite and Space Communications Technical Committee,  the chair of the IEEE special interest group entitled “Satellite Mega-constellations: Communications and Networking,” and also as an editor in 6 different IEEE journals. She is a member of the IEEE WCNC Steering Board and a Distinguished Lecturer of the Vehicular Technology Society Class of 2022. Her research interests include multi-functional space networks, space security, and wireless testbeds.
\end{biographywithpic}

\begin{biographywithpic}
{Antoine~Lesage-Landry}{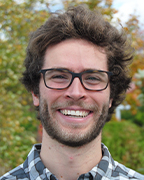}
(Senior~Member,~IEEE) is an Associate Professor in the Department of Electrical Engineering at Polytechnique Montréal, QC, Canada. He received the B.Eng. degree in Engineering Physics from Polytechnique Montréal, QC, Canada, in 2015, and the Ph.D. degree in Electrical Engineering from the University of Toronto, ON, Canada, in 2019. From 2019 to 2020, he was a Postdoctoral Scholar in the Energy \& Resources Group at the University of California, Berkeley, CA, USA. His research interests include optimization and machine learning, and their application to renewable power systems and wireless communications.
\end{biographywithpic}

\appendices{}\label{sec:app}             

\section{Inputs and Hyperparameters}\label{sec:hyperparameter_appendix}         
In this appendix, we provide detailed information about the input features and hyperparameter sweeps conducted for training the KAN- and PIKAN-based models. The selected input features include carrier frequency ($f_{\mathrm{c}}$), horizontal and vertical distances ($d_{\mathrm{hor}}, d_{\mathrm{ver}}$), and angles of arrival and departure ($\alpha_{\mathrm{AoA}}, \beta_{\mathrm{AoA}}, \alpha_{\mathrm{AoD}}, \beta_{\mathrm{AoD}}$), as detailed in~\SEC{sec:dataset}. The grid size $G$ and spline order $k$ are the key hyperparameters of the KAN. A grid search over $k~\in~\{2,3,4\}$ and $G~\in~\{5,10,15,20\}$ is performed to assess their effect on model performance. \TAB{tab:model_inputs_sweep} reports the test performance for different input-feature sets and hyperparameter combinations. The results highlight that the best-performing configuration is obtained with $k=3$ and $G=5$, providing a balanced trade-off between accuracy and generalization.

\section{Dataset}\label{sec:dataset_appendix}       
We provide detailed information about the dataset used for training and evaluation of the proposed KAN-based path loss model. \TAB{tab:meas_params} summarizes the measurement parameters adapted from~\cite{10757825}, including center frequencies, sampling rate, transmit power, and flight configurations. \FGR{fig:aerpaw_dataset} illustrates the variations of altitude, carrier frequency, angles of arrival and departure, distance, speed, path loss, and UAV orientation over time during the measurement campaigns. These details complement the main text by offering a comprehensive view of the experimental setup and the dynamic conditions captured in the dataset.

\begin{table}[!h]
\centering
\caption{Measurement Parameters (adapted from \cite{10757825}).}
\label{tab:meas_params}
\resizebox{\linewidth}{!}{%
\renewcommand{\arraystretch}{1.2}
\begin{tabular}{l l}
\hline\hline
\textbf{Parameter} & \textbf{Value} \\
\hline
Center Frequency & 3564, 3620, 3686 MHz \\
Sampling Rate & 56 MHz \\
Transmit Power & 19 dBm \\
Transmitted Waveform & Zadoff-Chu Sequence \\
Sequence Length & 401 \\
Root index & 200 \\
Repetition of Sequence & 4 \\
Measurement Frequency & 4 Hz \\
Altitude (above ground level) & 30, 60, 90 m \\
Flight Speed & 5 m/s \\
Flight Path & 500 m \\
\hline\hline
\end{tabular}}
\vspace{-0.2in}
\end{table}

\section{MLP Training}\label{sec:mlp_training}       
All MLP baselines in~\SEC{sec:nn_benchmark} are trained under a consistent experimental setup to ensure fair comparison. Each architecture employs ReLU activation functions, no dropout, and weight decay of $10^{-4}$ for regularization. The models are trained for up to 100 epochs with a batch size of 64, using the Adam optimizer with an initial learning rate of $10^{-3}$. The MSE is adopted as the loss function. To prevent overfitting, early stopping with a patience of 10 epochs is applied based on validation performance. The hidden-layer configurations for each MLP variant are provided in~\SEC{sec:nn_benchmark}. This uniform training strategy ensures that differences in performance stem from architectural capacity rather than optimization discrepancies.

\section{Additional Figures}\label{sec:additional_figs_appendix} 

This appendix provides additional scatter plots and comparisons to complement the main results in Section~V. Each figure illustrates both (a) path loss versus distance, and (b) measured versus predicted path loss, enabling a more detailed visual evaluation of the models’ performance. These extended plots are included here to avoid overcrowding the main body of the paper.

\onecolumn
\begin{center}
\small
\setlength{\tabcolsep}{6pt}
\renewcommand{\arraystretch}{1.2}
\setlength\LTleft{0pt}\setlength\LTright{0pt}

\begin{longtable}{*{12}{C{0.6cm}}cccc}
\caption{Test performance under different inputs and hyperparameters. The model with the lowest RMSE is highlighted with a shaded row and boxed RMSE/MAE values.}
\label{tab:model_inputs_sweep}\\

\hline\hline
\multicolumn{12}{c}{\textbf{Input Parameters}} &
\textbf{$k$} & \textbf{$G$} & \textbf{RMSE} & \textbf{MAE} \\
\cline{1-12}
$f_{\mathrm{c}}$ & $d$ & $h$ & $d_{\mathrm{hor}}$ & $d_{\mathrm{ver}}$ & ${\alpha}_{\text{AoA}}$ & ${\beta}_{\text{AoA}}$ & ${\alpha}_{\text{AoD}}$ & ${\beta}_{\text{AoD}}$ & $\phi$ & $\theta$ & $\psi$ & & & & \\
\hline
\endfirsthead

\multicolumn{16}{l}{\textit{Table \thetable\ (continued).}}\\[2pt]
\hline\hline
\multicolumn{12}{c}{\textbf{Input Parameters}} &
\textbf{$k$} & \textbf{$G$} & \textbf{RMSE} & \textbf{MAE} \\
\cline{1-12}
$f_{\mathrm{c}}$ & $d$ & $h$ & $d_{\mathrm{hor}}$ & $d_{\mathrm{ver}}$ & ${\alpha}_{\text{AoA}}$ & ${\beta}_{\text{AoA}}$ & ${\alpha}_{\text{AoD}}$ & ${\beta}_{\text{AoD}}$ & $\phi$ & $\theta$ & $\psi$ & & & & \\
\hline
\endhead

\hline
\multicolumn{16}{r}{\textit{Continued on next page}}\\
\hline
\endfoot

\hline\hline
\endlastfoot

\checkmark & \checkmark & \checkmark & $\times$ & $\times$ & $\times$ & $\times$ & $\times$ & $\times$ & $\times$ & $\times$ & $\times$ & 2 & 5 & 4.28 & 3.11 \\
\checkmark & \checkmark & \checkmark & $\times$ & $\times$ & $\times$ & $\times$ & $\times$ & $\times$ & $\times$ & $\times$ & $\times$ & 2 & 10 & 4.65 & 3.58 \\
\checkmark & \checkmark & \checkmark & $\times$ & $\times$ & $\times$ & $\times$ & $\times$ & $\times$ & $\times$ & $\times$ & $\times$ & 2 & 15 & 4.95 & 3.63 \\
\checkmark & \checkmark & \checkmark & $\times$ & $\times$ & $\times$ & $\times$ & $\times$ & $\times$ & $\times$ & $\times$ & $\times$ & 2 & 20 & 5.14 & 3.66 \\
\checkmark & \checkmark & \checkmark & $\times$ & $\times$ & $\times$ & $\times$ & $\times$ & $\times$ & $\times$ & $\times$ & $\times$ & 3 & 5 & 4.60 & 3.44 \\
\checkmark & \checkmark & \checkmark & $\times$ & $\times$ & $\times$ & $\times$ & $\times$ & $\times$ & $\times$ & $\times$ & $\times$ & 3 & 10 & 5.69 & 4.41 \\
\checkmark & \checkmark & \checkmark & $\times$ & $\times$ & $\times$ & $\times$ & $\times$ & $\times$ & $\times$ & $\times$ & $\times$ & 3 & 15 & 4.52 & 3.30 \\
\checkmark & \checkmark & \checkmark & $\times$ & $\times$ & $\times$ & $\times$ & $\times$ & $\times$ & $\times$ & $\times$ & $\times$ & 3 & 20 & 4.68 & 3.37 \\
\checkmark & \checkmark & \checkmark & $\times$ & $\times$ & $\times$ & $\times$ & $\times$ & $\times$ & $\times$ & $\times$ & $\times$ & 4 & 5 & 4.39 & 3.23 \\
\checkmark & \checkmark & \checkmark & $\times$ & $\times$ & $\times$ & $\times$ & $\times$ & $\times$ & $\times$ & $\times$ & $\times$ & 4 & 10 & 4.85 & 3.51 \\
\checkmark & \checkmark & \checkmark & $\times$ & $\times$ & $\times$ & $\times$ & $\times$ & $\times$ & $\times$ & $\times$ & $\times$ & 4 & 15 & 4.57 & 3.42 \\
\checkmark & \checkmark & \checkmark & $\times$ & $\times$ & $\times$ & $\times$ & $\times$ & $\times$ & $\times$ & $\times$ & $\times$ & 4 & 20 & 7.61 & 5.25 \\
\checkmark & \checkmark & \checkmark & \checkmark & \checkmark & $\times$ & $\times$ & $\times$ & $\times$ & $\times$ & $\times$ & $\times$ & 2 & 5 & 14.80 & 6.84 \\
\checkmark & \checkmark & \checkmark & \checkmark & \checkmark & $\times$ & $\times$ & $\times$ & $\times$ & $\times$ & $\times$ & $\times$ & 2 & 10 & 4.42 & 3.28 \\
\checkmark & \checkmark & \checkmark & \checkmark & \checkmark & $\times$ & $\times$ & $\times$ & $\times$ & $\times$ & $\times$ & $\times$ & 2 & 15 & 4.41 & 3.24 \\
\checkmark & \checkmark & \checkmark & \checkmark & \checkmark & $\times$ & $\times$ & $\times$ & $\times$ & $\times$ & $\times$ & $\times$ & 2 & 20 & 4.92 & 3.60 \\
\checkmark & \checkmark & \checkmark & \checkmark & \checkmark & $\times$ & $\times$ & $\times$ & $\times$ & $\times$ & $\times$ & $\times$ & 3 & 5 & 6.21 & 4.37 \\
\checkmark & \checkmark & \checkmark & \checkmark & \checkmark & $\times$ & $\times$ & $\times$ & $\times$ & $\times$ & $\times$ & $\times$ & 3 & 10 & 4.30 & 3.14 \\
\checkmark & \checkmark & \checkmark & \checkmark & \checkmark & $\times$ & $\times$ & $\times$ & $\times$ & $\times$ & $\times$ & $\times$ & 3 & 15 & 9.53 & 6.43 \\
\checkmark & \checkmark & \checkmark & \checkmark & \checkmark & $\times$ & $\times$ & $\times$ & $\times$ & $\times$ & $\times$ & $\times$ & 3 & 20 & 6.94 & 4.83 \\
\checkmark & \checkmark & \checkmark & \checkmark & \checkmark & $\times$ & $\times$ & $\times$ & $\times$ & $\times$ & $\times$ & $\times$ & 4 & 5 & 4.71 & 3.52 \\
\checkmark & \checkmark & \checkmark & \checkmark & \checkmark & $\times$ & $\times$ & $\times$ & $\times$ & $\times$ & $\times$ & $\times$ & 4 & 10 & 14.45 & 9.20 \\
\checkmark & \checkmark & \checkmark & \checkmark & \checkmark & $\times$ & $\times$ & $\times$ & $\times$ & $\times$ & $\times$ & $\times$ & 4 & 15 & 6.36 & 4.55 \\
\checkmark & \checkmark & \checkmark & \checkmark & \checkmark & $\times$ & $\times$ & $\times$ & $\times$ & $\times$ & $\times$ & $\times$ & 4 & 20 & 4.70 & 3.43 \\
\checkmark & $\times$ & $\times$ & \checkmark & \checkmark & $\times$ & $\times$ & $\times$ & $\times$ & $\times$ & $\times$ & $\times$ & 2 & 5 & 4.45 & 3.33 \\
\checkmark & $\times$ & $\times$ & \checkmark & \checkmark & $\times$ & $\times$ & $\times$ & $\times$ & $\times$ & $\times$ & $\times$ & 2 & 10 & 7.71 & 5.69 \\
\checkmark & $\times$ & $\times$ & \checkmark & \checkmark & $\times$ & $\times$ & $\times$ & $\times$ & $\times$ & $\times$ & $\times$ & 2 & 15 & 4.67 & 3.40 \\
\checkmark & $\times$ & $\times$ & \checkmark & \checkmark & $\times$ & $\times$ & $\times$ & $\times$ & $\times$ & $\times$ & $\times$ & 2 & 20 & 6.86 & 4.88 \\
\checkmark & $\times$ & $\times$ & \checkmark & \checkmark & $\times$ & $\times$ & $\times$ & $\times$ & $\times$ & $\times$ & $\times$ & 3 & 5 & 4.58 & 3.30 \\
\checkmark & $\times$ & $\times$ & \checkmark & \checkmark & $\times$ & $\times$ & $\times$ & $\times$ & $\times$ & $\times$ & $\times$ & 3 & 10 & 4.55 & 3.23 \\
\checkmark & $\times$ & $\times$ & \checkmark & \checkmark & $\times$ & $\times$ & $\times$ & $\times$ & $\times$ & $\times$ & $\times$ & 3 & 15 & 5.70 & 3.97 \\
\checkmark & $\times$ & $\times$ & \checkmark & \checkmark & $\times$ & $\times$ & $\times$ & $\times$ & $\times$ & $\times$ & $\times$ & 3 & 20 & 4.39 & 3.12 \\
\checkmark & $\times$ & $\times$ & \checkmark & \checkmark & $\times$ & $\times$ & $\times$ & $\times$ & $\times$ & $\times$ & $\times$ & 4 & 5 & 4.40 & 3.17 \\
\checkmark & $\times$ & $\times$ & \checkmark & \checkmark & $\times$ & $\times$ & $\times$ & $\times$ & $\times$ & $\times$ & $\times$ & 4 & 10 & 4.57 & 3.41 \\
\checkmark & $\times$ & $\times$ & \checkmark & \checkmark & $\times$ & $\times$ & $\times$ & $\times$ & $\times$ & $\times$ & $\times$ & 4 & 15 & 7.19 & 5.11 \\
\checkmark & $\times$ & $\times$ & \checkmark & \checkmark & $\times$ & $\times$ & $\times$ & $\times$ & $\times$ & $\times$ & $\times$ & 4 & 20 & 5.74 & 4.30 \\
\checkmark & $\times$ & $\times$ & \checkmark & \checkmark & \checkmark & \checkmark & $\times$ & $\times$ & $\times$ & $\times$ & $\times$ & 2 & 5 & 4.37 & 3.15 \\
\checkmark & $\times$ & $\times$ & \checkmark & \checkmark & \checkmark & \checkmark & $\times$ & $\times$ & $\times$ & $\times$ & $\times$ & 2 & 10 & 4.37 & 3.17 \\
\checkmark & $\times$ & $\times$ & \checkmark & \checkmark & \checkmark & \checkmark & $\times$ & $\times$ & $\times$ & $\times$ & $\times$ & 2 & 15 & 4.34 & 3.14 \\
\checkmark & $\times$ & $\times$ & \checkmark & \checkmark & \checkmark & \checkmark & $\times$ & $\times$ & $\times$ & $\times$ & $\times$ & 2 & 20 & 4.92 & 3.52 \\
\checkmark & $\times$ & $\times$ & \checkmark & \checkmark & \checkmark & \checkmark & $\times$ & $\times$ & $\times$ & $\times$ & $\times$ & 3 & 5 & 5.69 & 4.24 \\
\checkmark & $\times$ & $\times$ & \checkmark & \checkmark & \checkmark & \checkmark & $\times$ & $\times$ & $\times$ & $\times$ & $\times$ & 3 & 10 & 5.17 & 3.68 \\
\checkmark & $\times$ & $\times$ & \checkmark & \checkmark & \checkmark & \checkmark & $\times$ & $\times$ & $\times$ & $\times$ & $\times$ & 3 & 15 & 4.81 & 3.31 \\
\checkmark & $\times$ & $\times$ & \checkmark & \checkmark & \checkmark & \checkmark & $\times$ & $\times$ & $\times$ & $\times$ & $\times$ & 3 & 20 & 6.49 & 4.78 \\
\checkmark & $\times$ & $\times$ & \checkmark & \checkmark & \checkmark & \checkmark & $\times$ & $\times$ & $\times$ & $\times$ & $\times$ & 4 & 5 & 7.65 & 4.90 \\
\checkmark & $\times$ & $\times$ & \checkmark & \checkmark & \checkmark & \checkmark & $\times$ & $\times$ & $\times$ & $\times$ & $\times$ & 4 & 10 & 4.25 & 3.05 \\
\checkmark & $\times$ & $\times$ & \checkmark & \checkmark & \checkmark & \checkmark & $\times$ & $\times$ & $\times$ & $\times$ & $\times$ & 4 & 15 & 12.25 & 6.78 \\
\checkmark & $\times$ & $\times$ & \checkmark & \checkmark & \checkmark & \checkmark & $\times$ & $\times$ & $\times$ & $\times$ & $\times$ & 4 & 20 & 6.50 & 4.72 \\
\checkmark & $\times$ & $\times$ & \checkmark & \checkmark & \checkmark & \checkmark & \checkmark & \checkmark & $\times$ & $\times$ & $\times$ & 2 & 5 & 4.47 & 3.23 \\
\checkmark & $\times$ & $\times$ & \checkmark & \checkmark & \checkmark & \checkmark & \checkmark & \checkmark & $\times$ & $\times$ & $\times$ & 2 & 10 & 4.14 & 3.22 \\
\checkmark & $\times$ & $\times$ & \checkmark & \checkmark & \checkmark & \checkmark & \checkmark & \checkmark & $\times$ & $\times$ & $\times$ & 2 & 15 & 4.64 & 3.30 \\
\checkmark & $\times$ & $\times$ & \checkmark & \checkmark & \checkmark & \checkmark & \checkmark & \checkmark & $\times$ & $\times$ & $\times$ & 2 & 20 & 4.68 & 3.33 \\
\rowcolor{gray!20}
\checkmark & $\times$ & $\times$ & \checkmark & \checkmark & \checkmark & \checkmark & \checkmark & \checkmark & $\times$ & $\times$ & $\times$ & 3 & 5 & \textbf{3.98} & \textbf{3.08} \\
\checkmark & $\times$ & $\times$ & \checkmark & \checkmark & \checkmark & \checkmark & \checkmark & \checkmark & $\times$ & $\times$ & $\times$ & 3 & 10 & 4.44 & 3.23 \\
\checkmark & $\times$ & $\times$ & \checkmark & \checkmark & \checkmark & \checkmark & \checkmark & \checkmark & $\times$ & $\times$ & $\times$ & 3 & 15 & 4.69 & 3.36 \\
\checkmark & $\times$ & $\times$ & \checkmark & \checkmark & \checkmark & \checkmark & \checkmark & \checkmark & $\times$ & $\times$ & $\times$ & 3 & 20 & 4.74 & 3.33 \\
\checkmark & $\times$ & $\times$ & \checkmark & \checkmark & \checkmark & \checkmark & \checkmark & \checkmark & $\times$ & $\times$ & $\times$ & 4 & 5 & 6.14 & 4.44 \\
\checkmark & $\times$ & $\times$ & \checkmark & \checkmark & \checkmark & \checkmark & \checkmark & \checkmark & $\times$ & $\times$ & $\times$ & 4 & 10 & 4.28 & 3.07 \\
\checkmark & $\times$ & $\times$ & \checkmark & \checkmark & \checkmark & \checkmark & \checkmark & \checkmark & $\times$ & $\times$ & $\times$ & 4 & 15 & 4.85 & 3.54 \\
\checkmark & $\times$ & $\times$ & \checkmark & \checkmark & \checkmark & \checkmark & \checkmark & \checkmark & $\times$ & $\times$ & $\times$ & 4 & 20 & 5.33 & 3.81 \\
\checkmark & $\times$ & $\times$ & \checkmark & \checkmark & \checkmark & \checkmark & \checkmark & \checkmark & \checkmark & \checkmark & \checkmark & 2 & 5 & 4.24 & 3.15 \\
\checkmark & $\times$ & $\times$ & \checkmark & \checkmark & \checkmark & \checkmark & \checkmark & \checkmark & \checkmark & \checkmark & \checkmark & 2 & 10 & 4.56 & 3.27 \\
\checkmark & $\times$ & $\times$ & \checkmark & \checkmark & \checkmark & \checkmark & \checkmark & \checkmark & \checkmark & \checkmark & \checkmark & 2 & 15 & 4.64 & 3.42 \\
\checkmark & $\times$ & $\times$ & \checkmark & \checkmark & \checkmark & \checkmark & \checkmark & \checkmark & \checkmark & \checkmark & \checkmark & 2 & 20 & 4.82 & 3.41 \\
\checkmark & $\times$ & $\times$ & \checkmark & \checkmark & \checkmark & \checkmark & \checkmark & \checkmark & \checkmark & \checkmark & \checkmark & 3 & 5 & 4.27 & 3.20 \\
\checkmark & $\times$ & $\times$ & \checkmark & \checkmark & \checkmark & \checkmark & \checkmark & \checkmark & \checkmark & \checkmark & \checkmark & 3 & 10 & 4.33 & 3.06 \\
\checkmark & $\times$ & $\times$ & \checkmark & \checkmark & \checkmark & \checkmark & \checkmark & \checkmark & \checkmark & \checkmark & \checkmark & 3 & 15 & 4.95 & 3.78 \\
\checkmark & $\times$ & $\times$ & \checkmark & \checkmark & \checkmark & \checkmark & \checkmark & \checkmark & \checkmark & \checkmark & \checkmark & 3 & 20 & 4.76 & 3.31 \\
\checkmark & $\times$ & $\times$ & \checkmark & \checkmark & \checkmark & \checkmark & \checkmark & \checkmark & \checkmark & \checkmark & \checkmark & 4 & 5 & 11.83 & 8.12 \\
\checkmark & $\times$ & $\times$ & \checkmark & \checkmark & \checkmark & \checkmark & \checkmark & \checkmark & \checkmark & \checkmark & \checkmark & 4 & 10 & 5.56 & 3.80 \\
\checkmark & $\times$ & $\times$ & \checkmark & \checkmark & \checkmark & \checkmark & \checkmark & \checkmark & \checkmark & \checkmark & \checkmark & 4 & 15 & 4.90 & 3.44 \\
\checkmark & $\times$ & $\times$ & \checkmark & \checkmark & \checkmark & \checkmark & \checkmark & \checkmark & \checkmark & \checkmark & \checkmark & 4 & 20 & 4.49 & 3.14 \\
\checkmark & $\times$ & $\times$ & \checkmark & \checkmark & $\times$ & $\times$ & $\times$ & $\times$ & \checkmark & \checkmark & \checkmark & 2 & 5 & 4.63 & 3.35 \\
\checkmark & $\times$ & $\times$ & \checkmark & \checkmark & $\times$ & $\times$ & $\times$ & $\times$ & \checkmark & \checkmark & \checkmark & 2 & 10 & 4.95 & 3.48 \\
\checkmark & $\times$ & $\times$ & \checkmark & \checkmark & $\times$ & $\times$ & $\times$ & $\times$ & \checkmark & \checkmark & \checkmark & 2 & 15 & 5.01 & 3.54 \\
\checkmark & $\times$ & $\times$ & \checkmark & \checkmark & $\times$ & $\times$ & $\times$ & $\times$ & \checkmark & \checkmark & \checkmark & 2 & 20 & 5.50 & 4.03 \\
\checkmark & $\times$ & $\times$ & \checkmark & \checkmark & $\times$ & $\times$ & $\times$ & $\times$ & \checkmark & \checkmark & \checkmark & 3 & 5 & 4.90 & 3.50 \\
\checkmark & $\times$ & $\times$ & \checkmark & \checkmark & $\times$ & $\times$ & $\times$ & $\times$ & \checkmark & \checkmark & \checkmark & 3 & 10 & 6.52 & 4.80 \\
\checkmark & $\times$ & $\times$ & \checkmark & \checkmark & $\times$ & $\times$ & $\times$ & $\times$ & \checkmark & \checkmark & \checkmark & 3 & 15 & 5.91 & 4.15 \\
\checkmark & $\times$ & $\times$ & \checkmark & \checkmark & $\times$ & $\times$ & $\times$ & $\times$ & \checkmark & \checkmark & \checkmark & 3 & 20 & 6.85 & 4.99 \\
\checkmark & $\times$ & $\times$ & \checkmark & \checkmark & $\times$ & $\times$ & $\times$ & $\times$ & \checkmark & \checkmark & \checkmark & 4 & 5 & 4.81 & 3.65 \\
\checkmark & $\times$ & $\times$ & \checkmark & \checkmark & $\times$ & $\times$ & $\times$ & $\times$ & \checkmark & \checkmark & \checkmark & 4 & 10 & 6.03 & 4.30 \\
\checkmark & $\times$ & $\times$ & \checkmark & \checkmark & $\times$ & $\times$ & $\times$ & $\times$ & \checkmark & \checkmark & \checkmark & 4 & 15 & 5.24 & 3.62 \\
\checkmark & $\times$ & $\times$ & \checkmark & \checkmark & $\times$ & $\times$ & $\times$ & $\times$ & \checkmark & \checkmark & \checkmark & 4 & 20 & 7.76 & 5.79 \\

\end{longtable}
\end{center}
\twocolumn

\begin{figure*}[!t]
    \centering
    \subfigure[]{
        \label{fig:aerpaw_alt_vs_time}
        \includegraphics[width=0.3\linewidth]{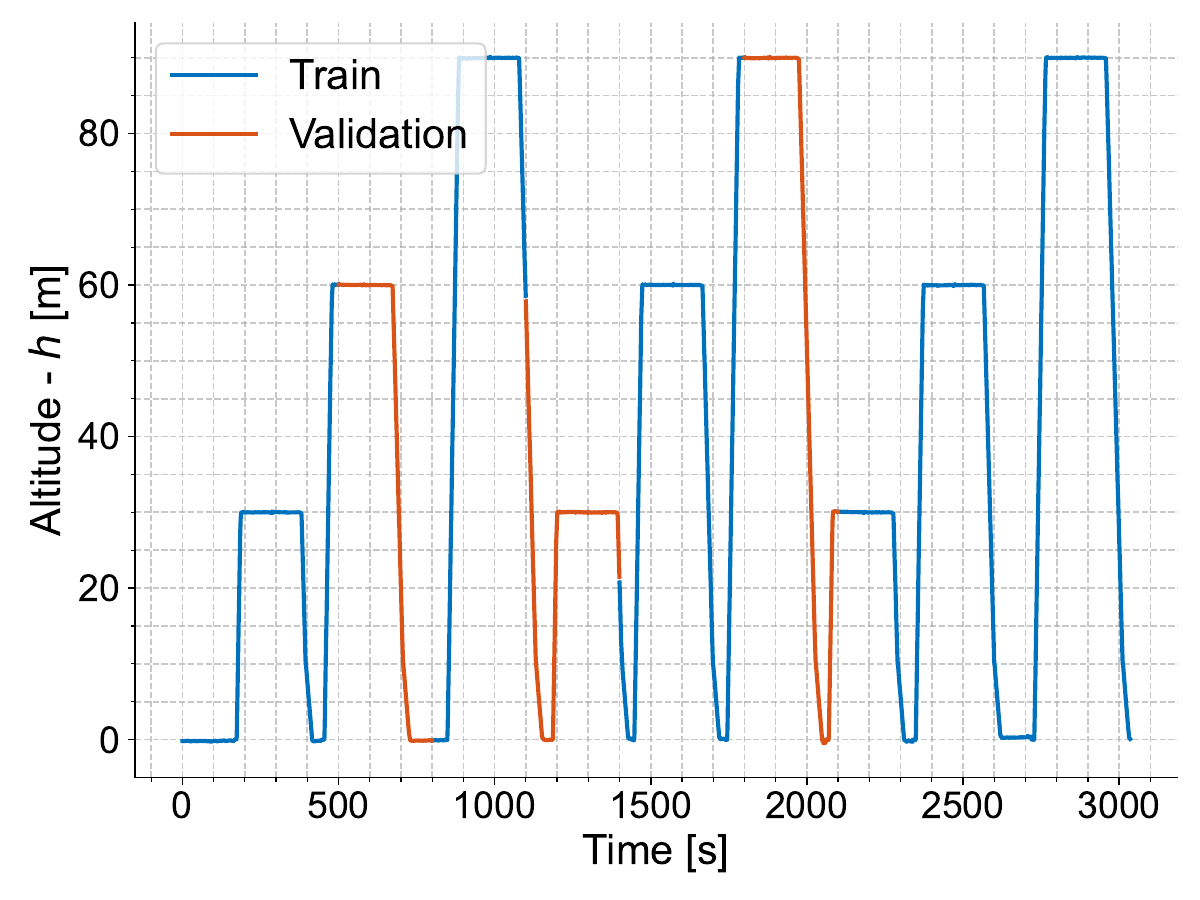}}
    \quad
    \subfigure[]{
        \label{fig:aerpaw_center_freq_vs_time}
        \includegraphics[width=0.3\linewidth]{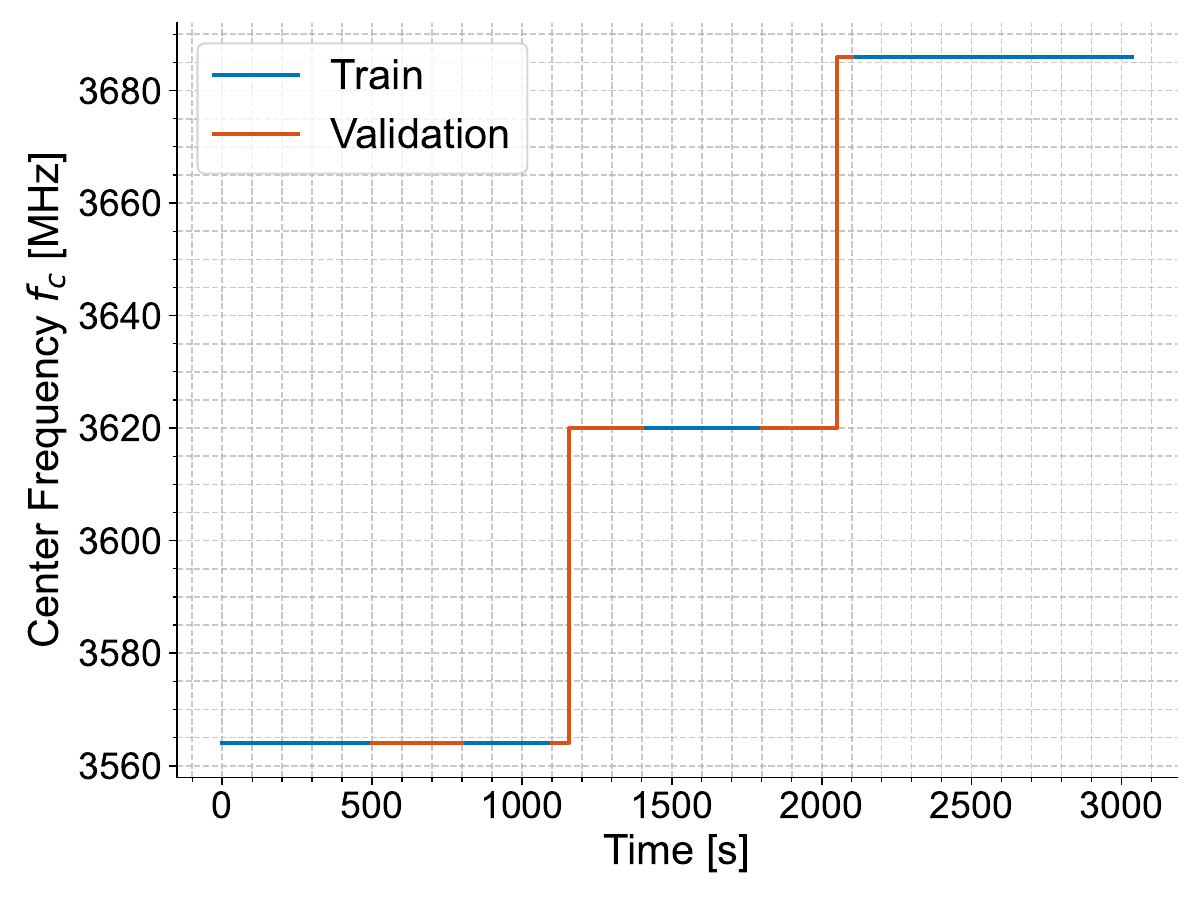}}
        \quad
    \subfigure[]{
        \label{fig:aerpaw_aoa_phi_vs_time}
        \includegraphics[width=0.3\linewidth]{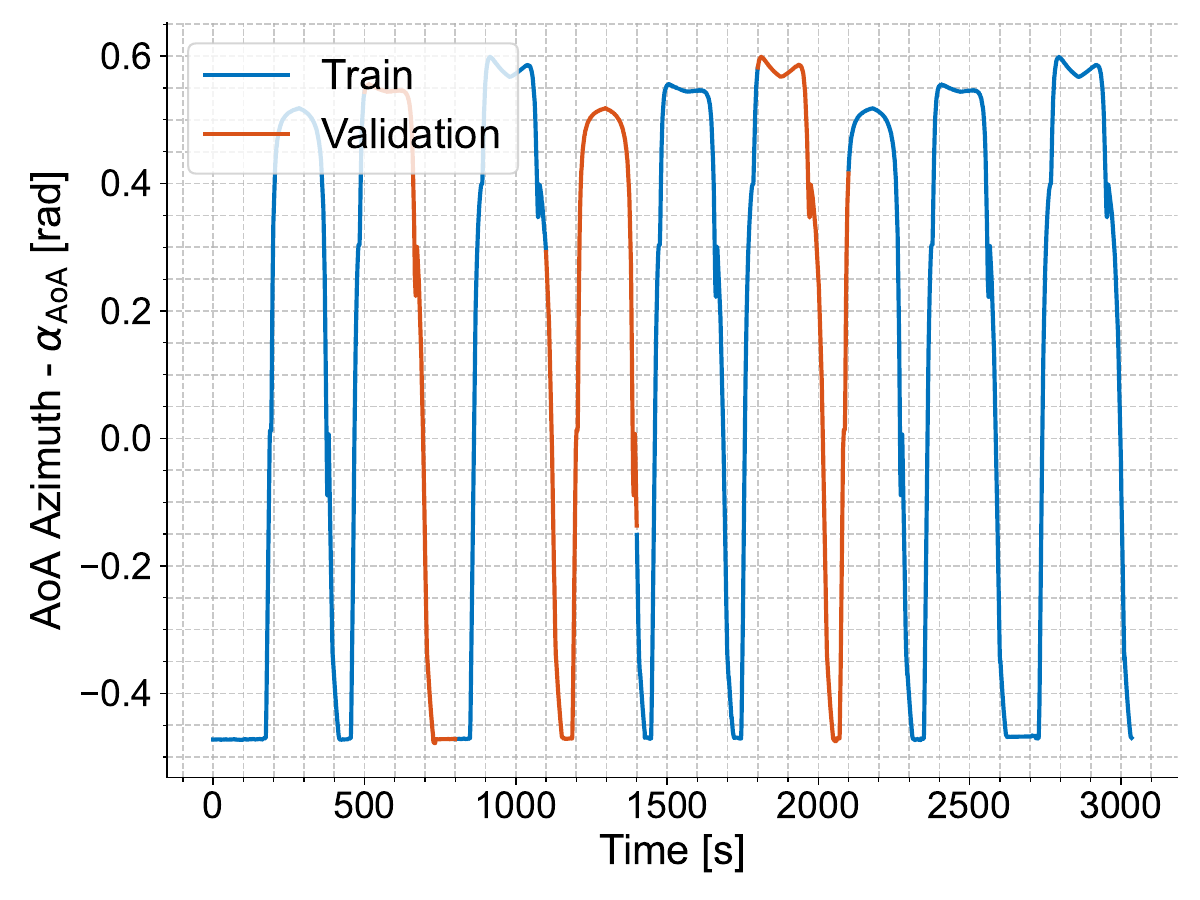}}
    \quad
    \subfigure[]{
        \label{fig:aerpaw_aoa_theta_vs_time}
        \includegraphics[width=0.3\linewidth]{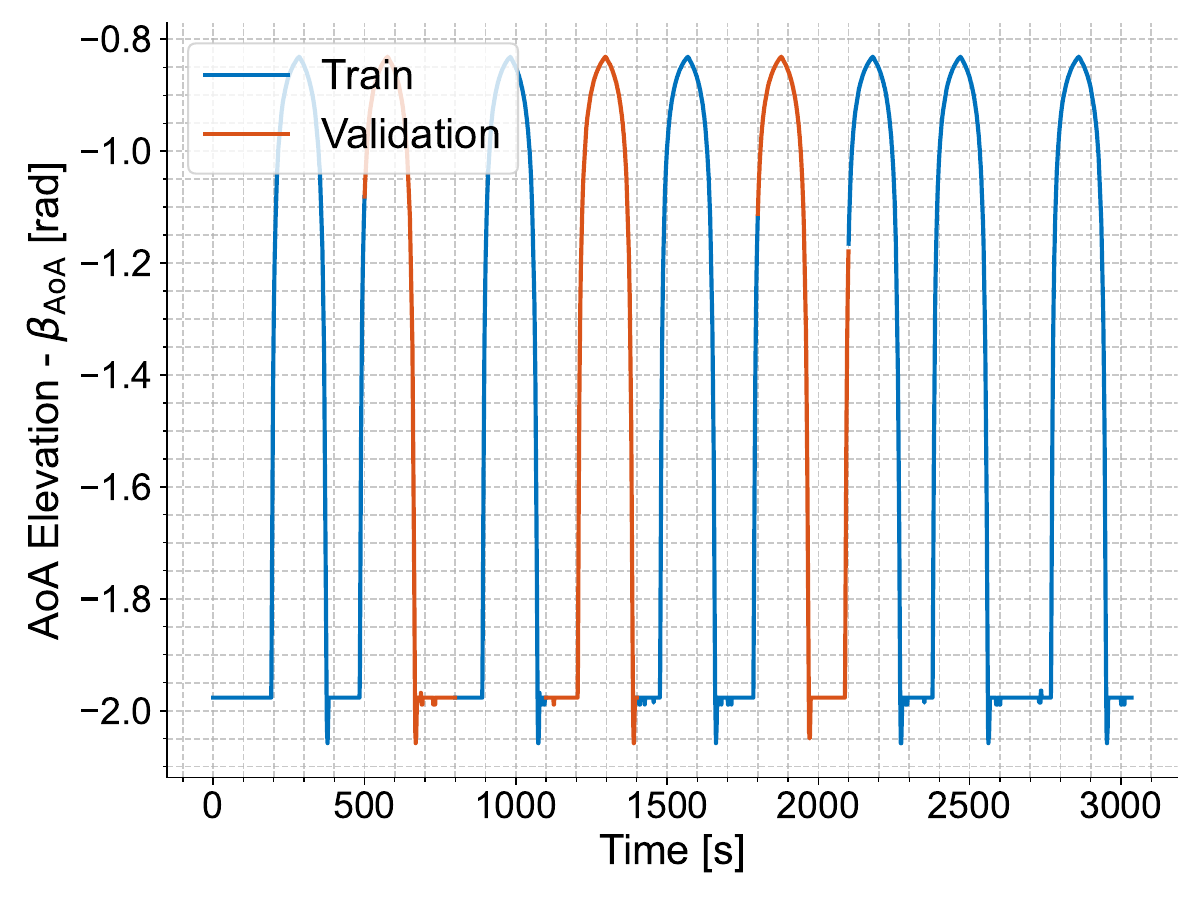}}
    \subfigure[]{
        \label{fig:aerpaw_aod_phi_vs_time}
        \includegraphics[width=0.3\linewidth]{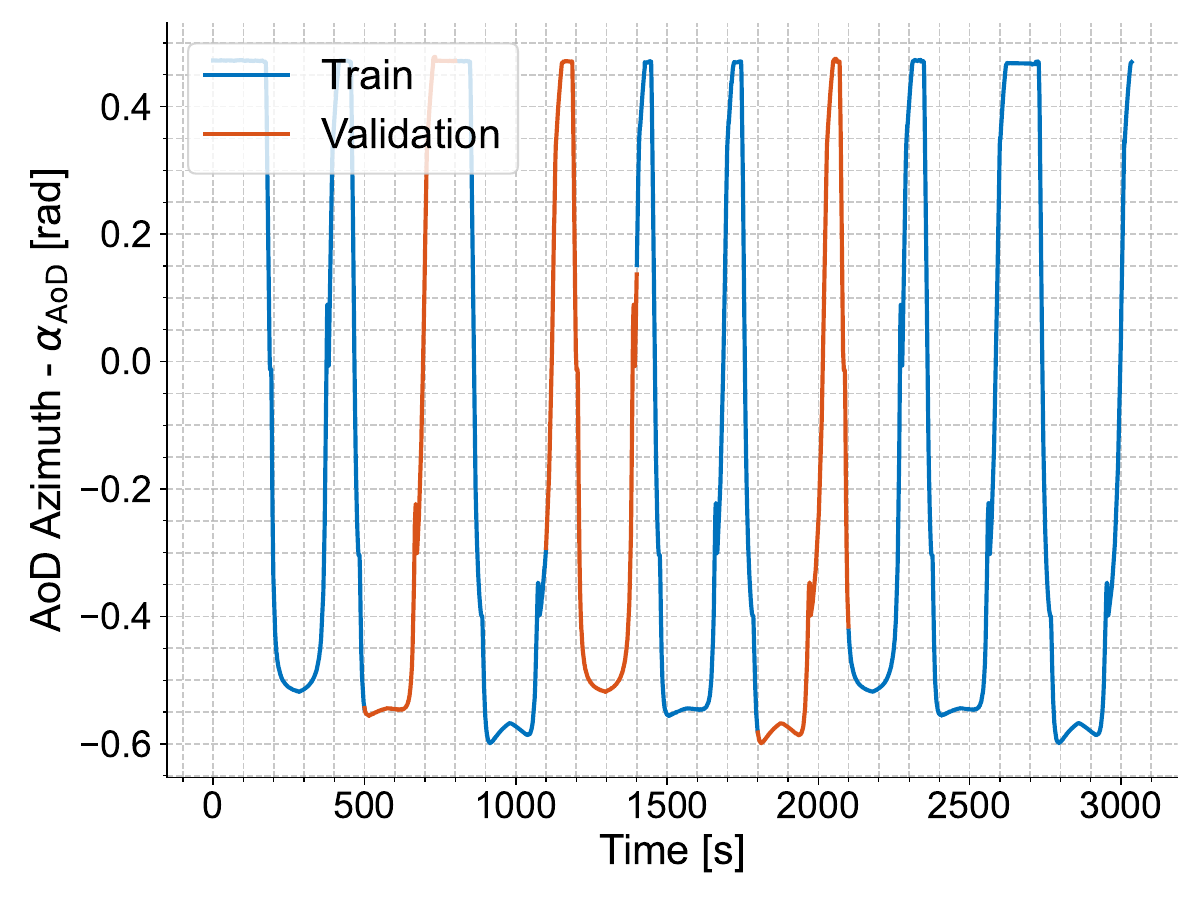}}
    \quad
    \subfigure[]{
        \label{fig:aerpaw_aod_theta_vs_time}
        \includegraphics[width=0.3\linewidth]{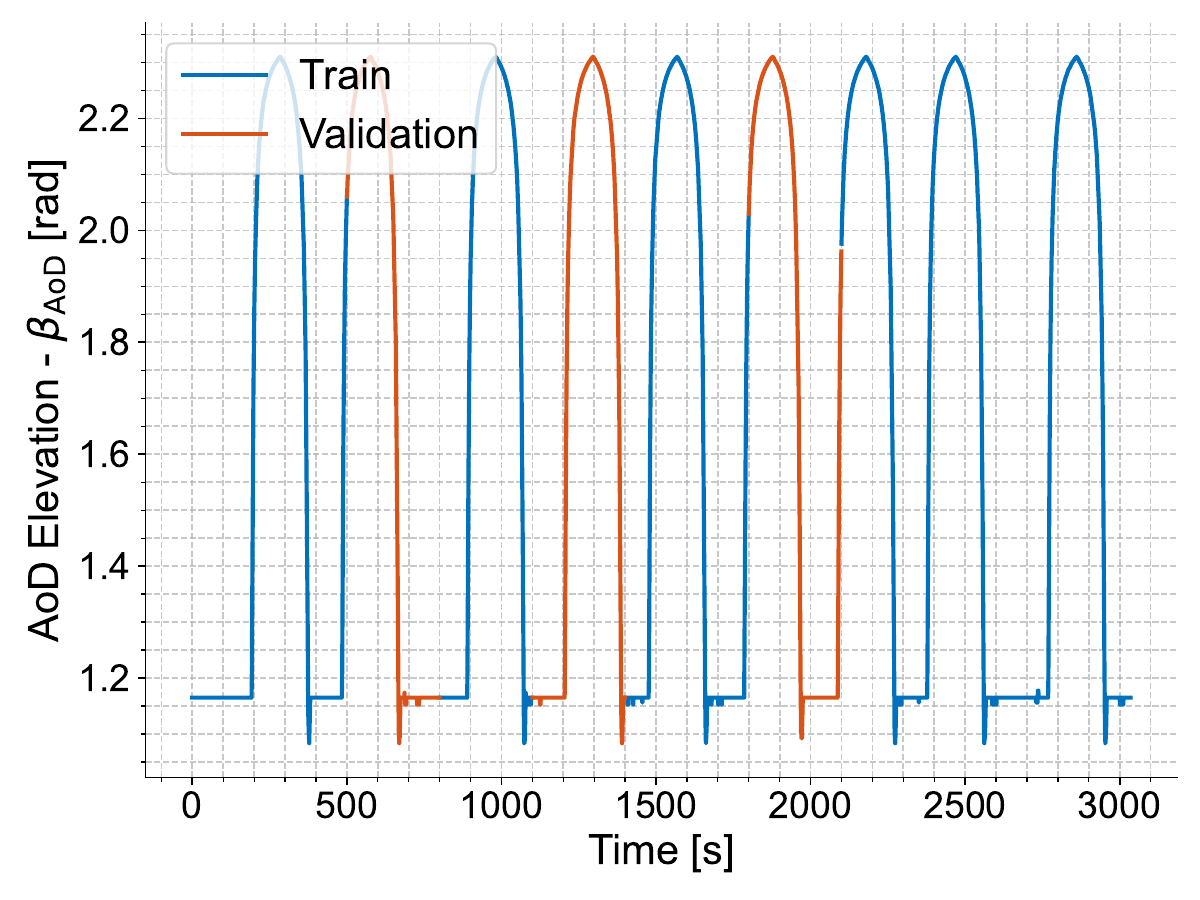}}
        \quad
    \subfigure[]{
        \label{fig:aerpaw_dist_vs_time}
        \includegraphics[width=0.3\linewidth]{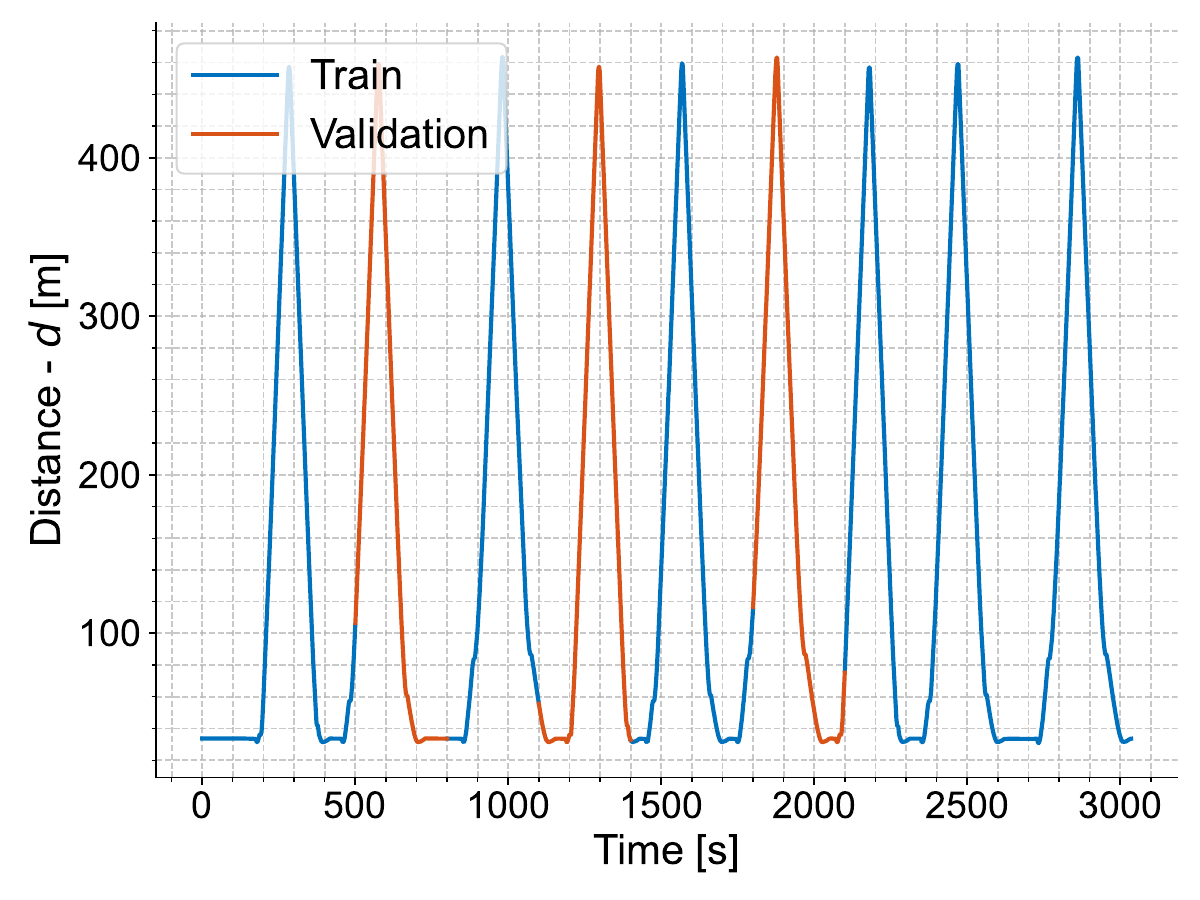}}
    \quad
    \subfigure[]{
        \label{fig:aerpaw_speed_vs_time}
        \includegraphics[width=0.3\linewidth]{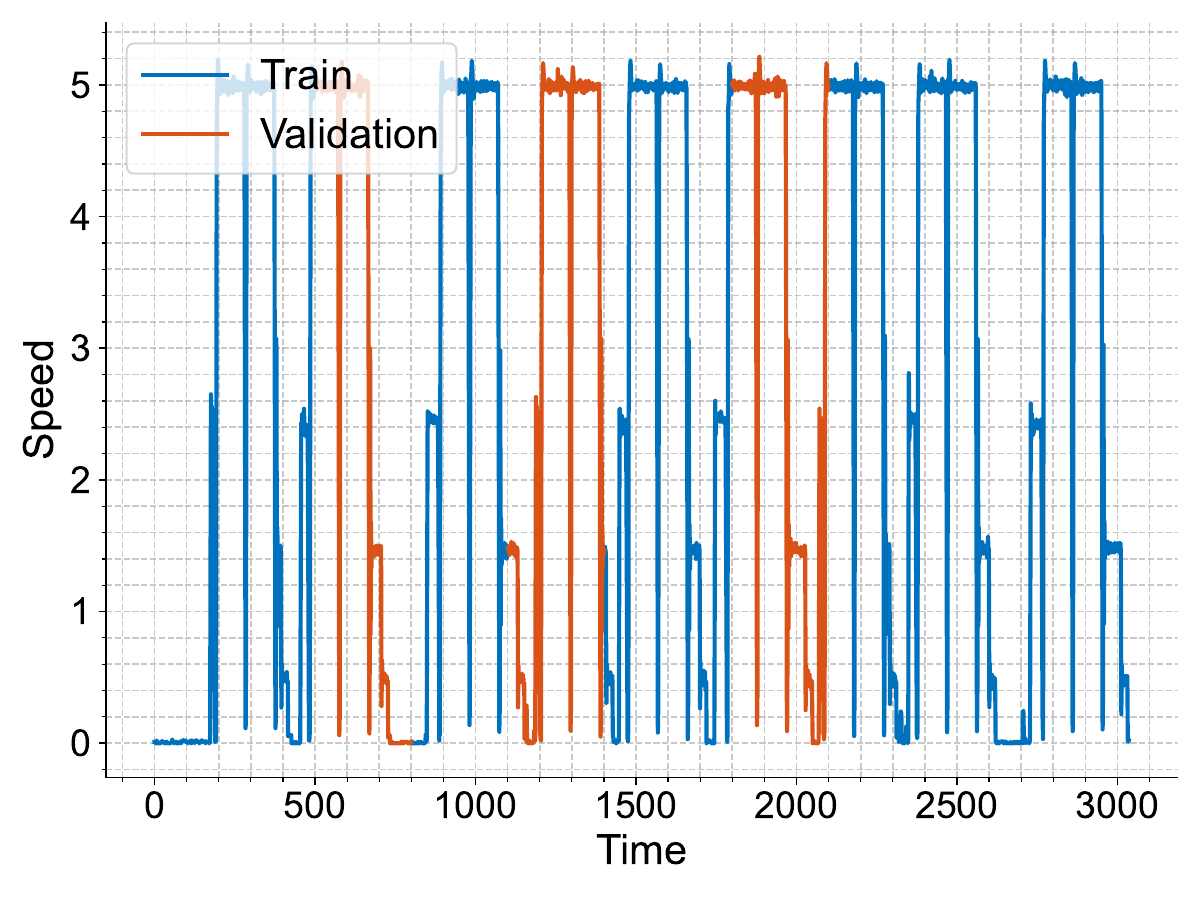}}
    \subfigure[]{
        \label{fig:aerpaw_avg_pl_vs_time}
        \includegraphics[width=0.3\linewidth]{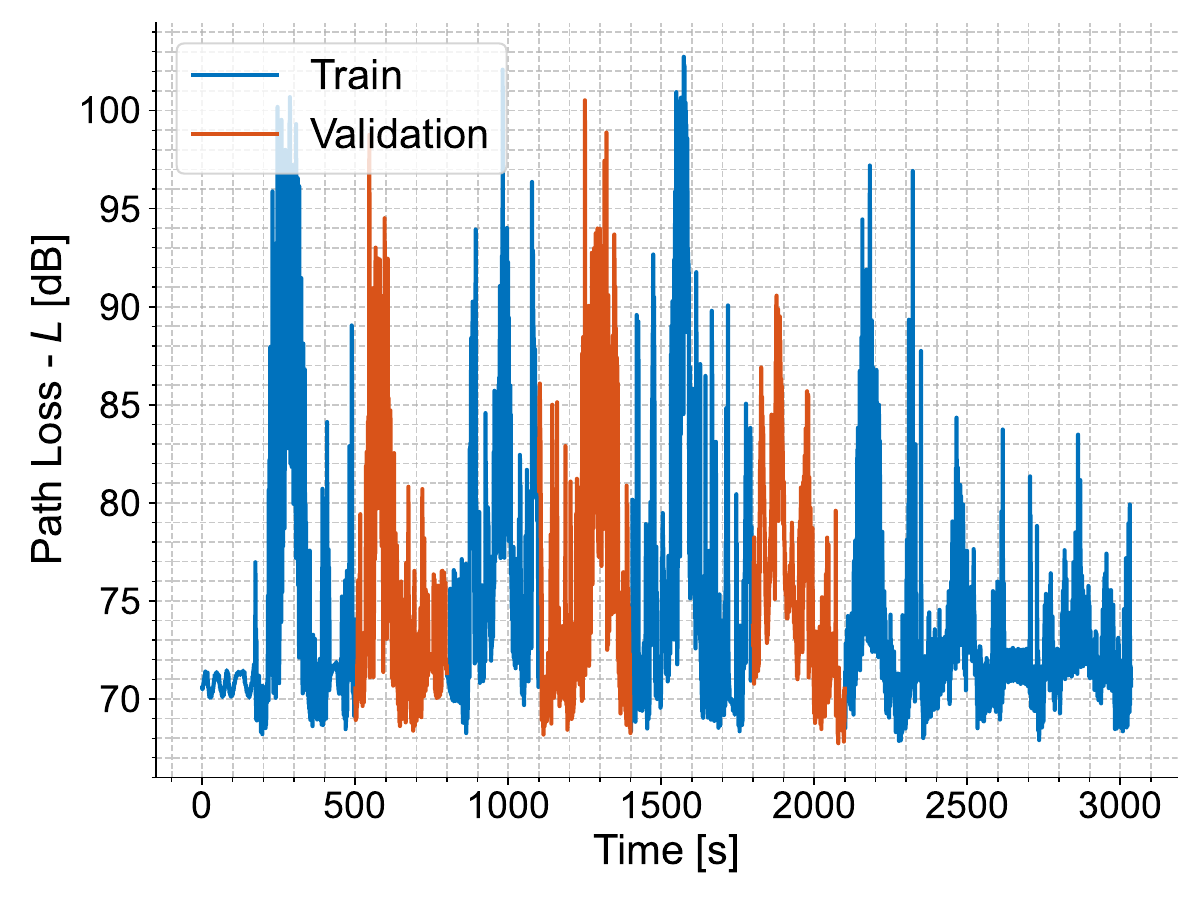}}
    \quad
    \subfigure[]{
        \label{fig:aerpaw_yaw_vs_time}
        \includegraphics[width=0.3\linewidth]{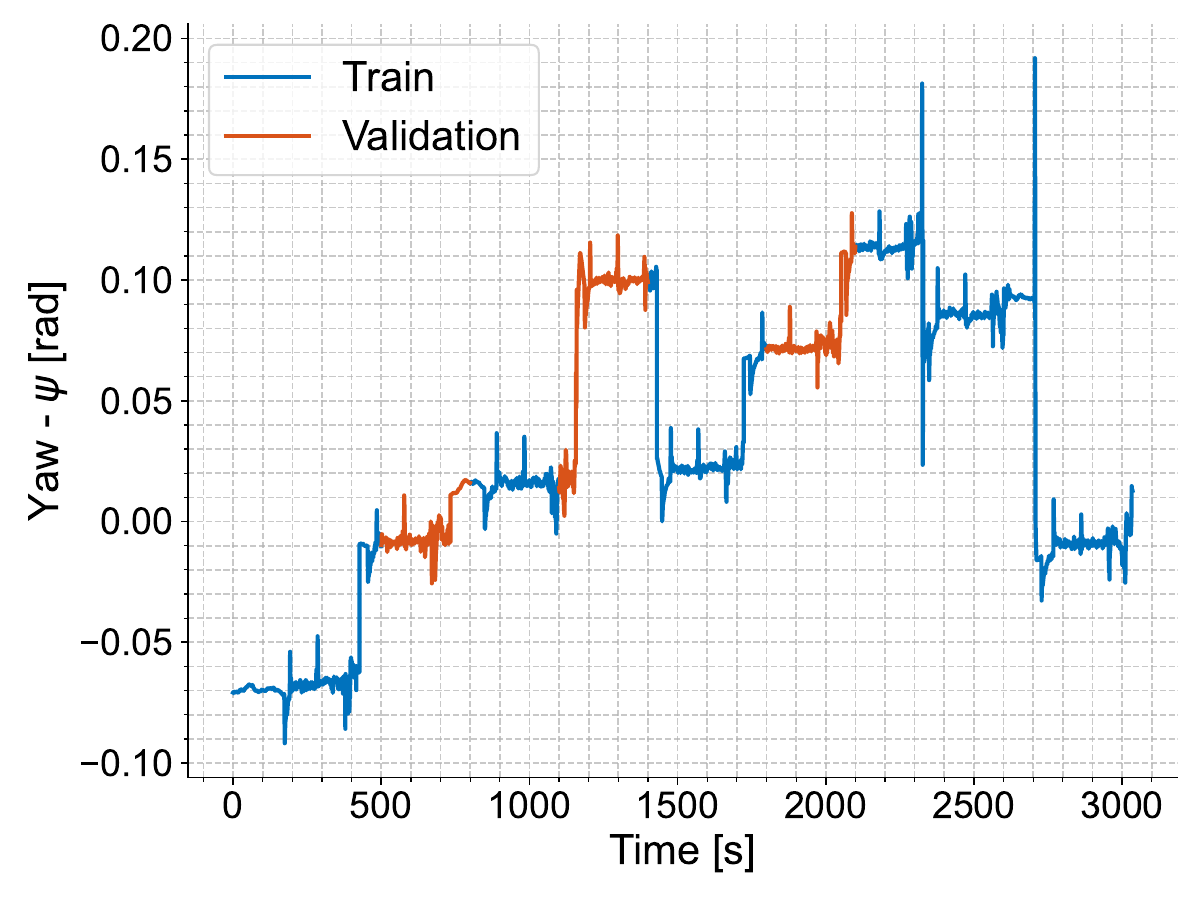}}
        \quad
    \subfigure[]{
        \label{fig:aerpaw_roll_vs_time}
        \includegraphics[width=0.3\linewidth]{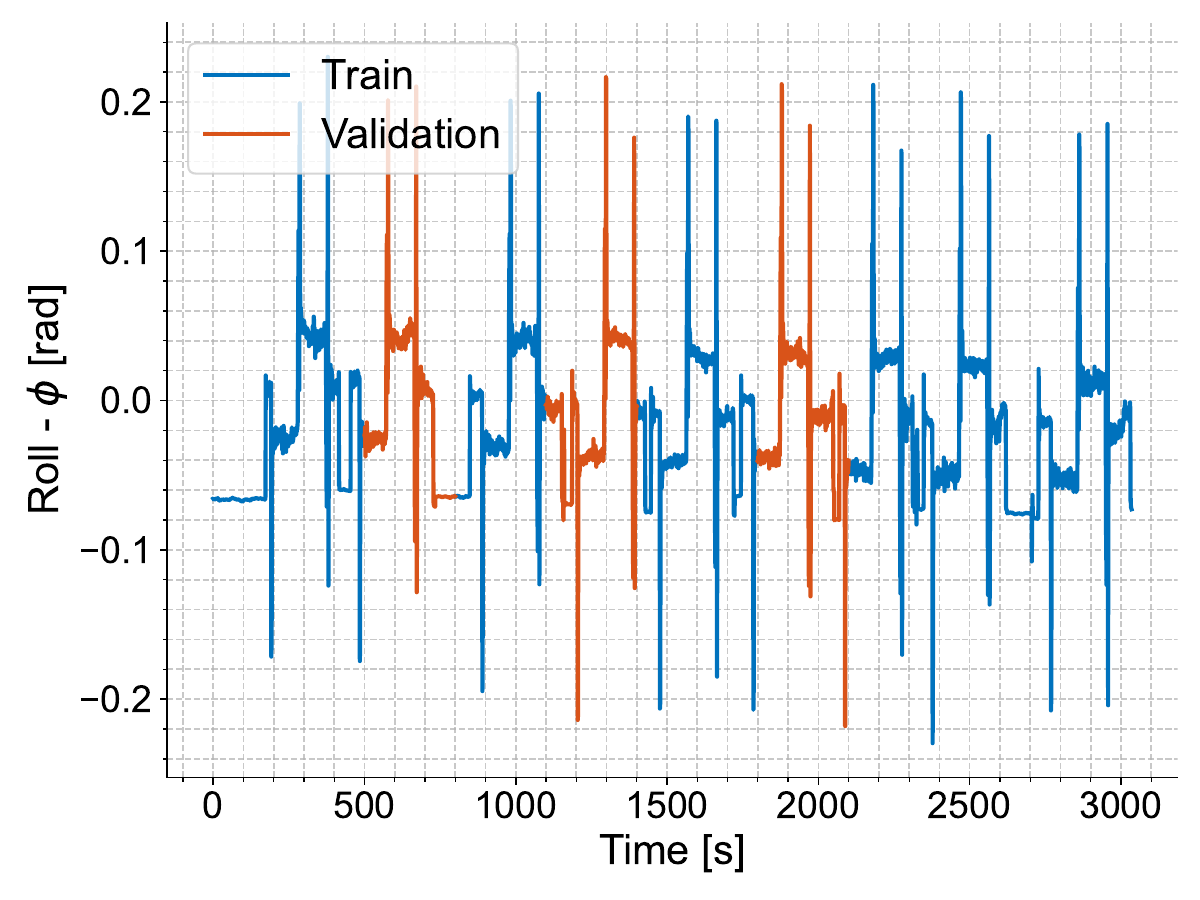}}
    \quad
    \subfigure[]{
        \label{fig:aerpaw_pitch_vs_time}
        \includegraphics[width=0.3\linewidth]{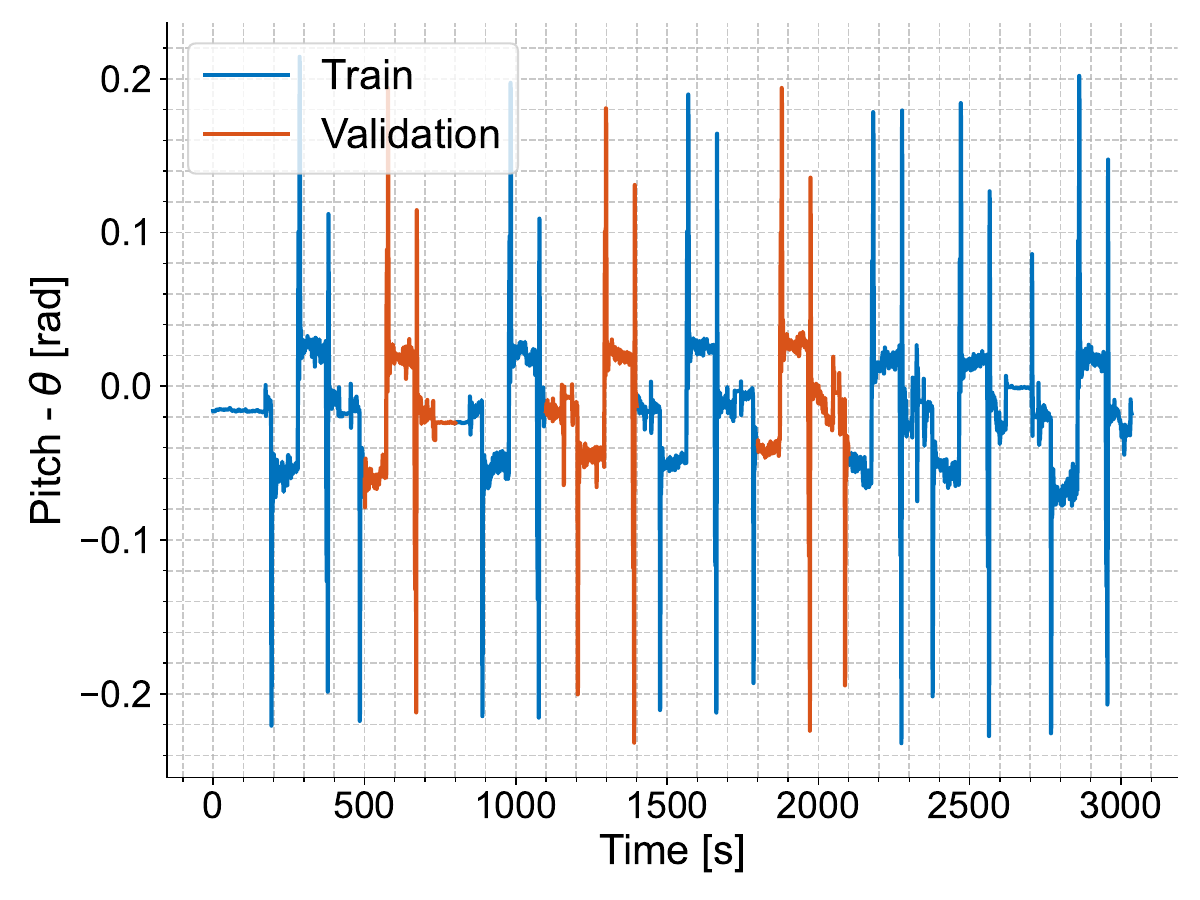}}
    \caption{Dataset characteristics over time during UAV measurement campaigns: 
            (a) UAV altitude, 
            (b) center frequency, 
            (c) AoA azimuth, 
            (d) AoA elevation, 
            (e) AoD azimuth, 
            (f) AoD elevation,
            (g) distance, 
            (h) UAV speed, 
            (i) measured path loss, 
            (j) UAV yaw, 
            (k) UAV roll, 
            (l) UAV pitch.}
    \label{fig:aerpaw_dataset}
\end{figure*}

\begin{figure*}[!t]
    \centering
    \subfigure[Base KAN: distance vs. path loss]{
        \includegraphics[width=0.22\linewidth]{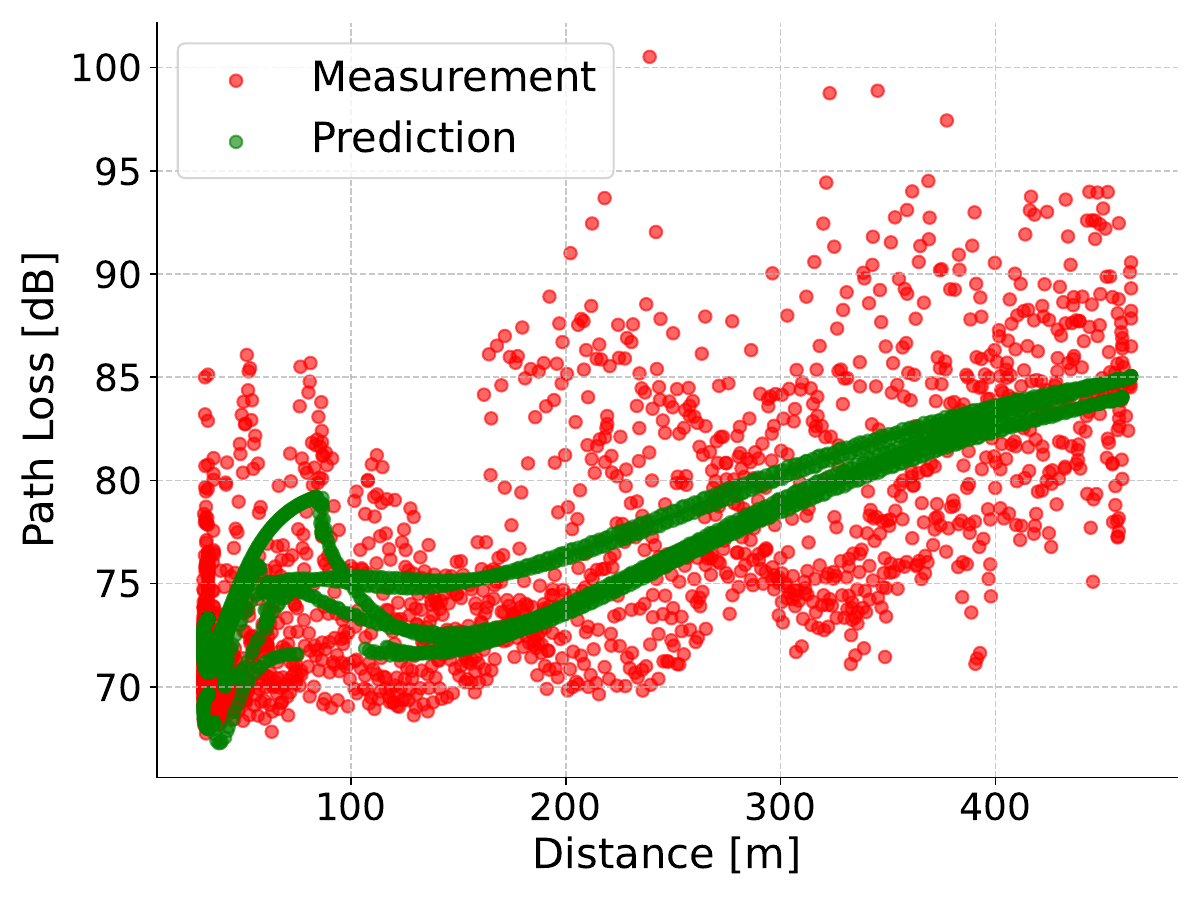}}
    \subfigure[Base KAN: measured vs. predicted]{
        \includegraphics[width=0.22\linewidth]{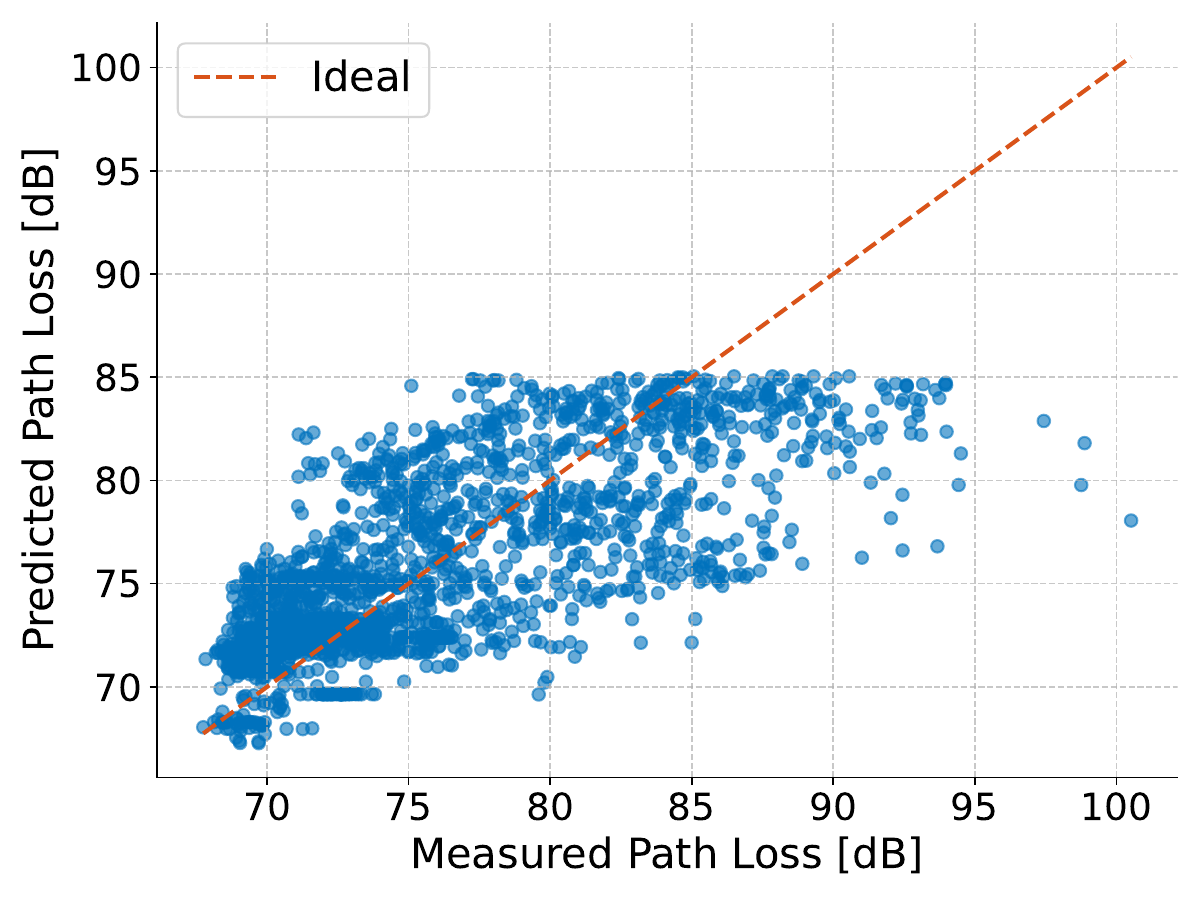}}
    \subfigure[Symbolic Base KAN: distance vs. path loss]{
        \includegraphics[width=0.22\linewidth]{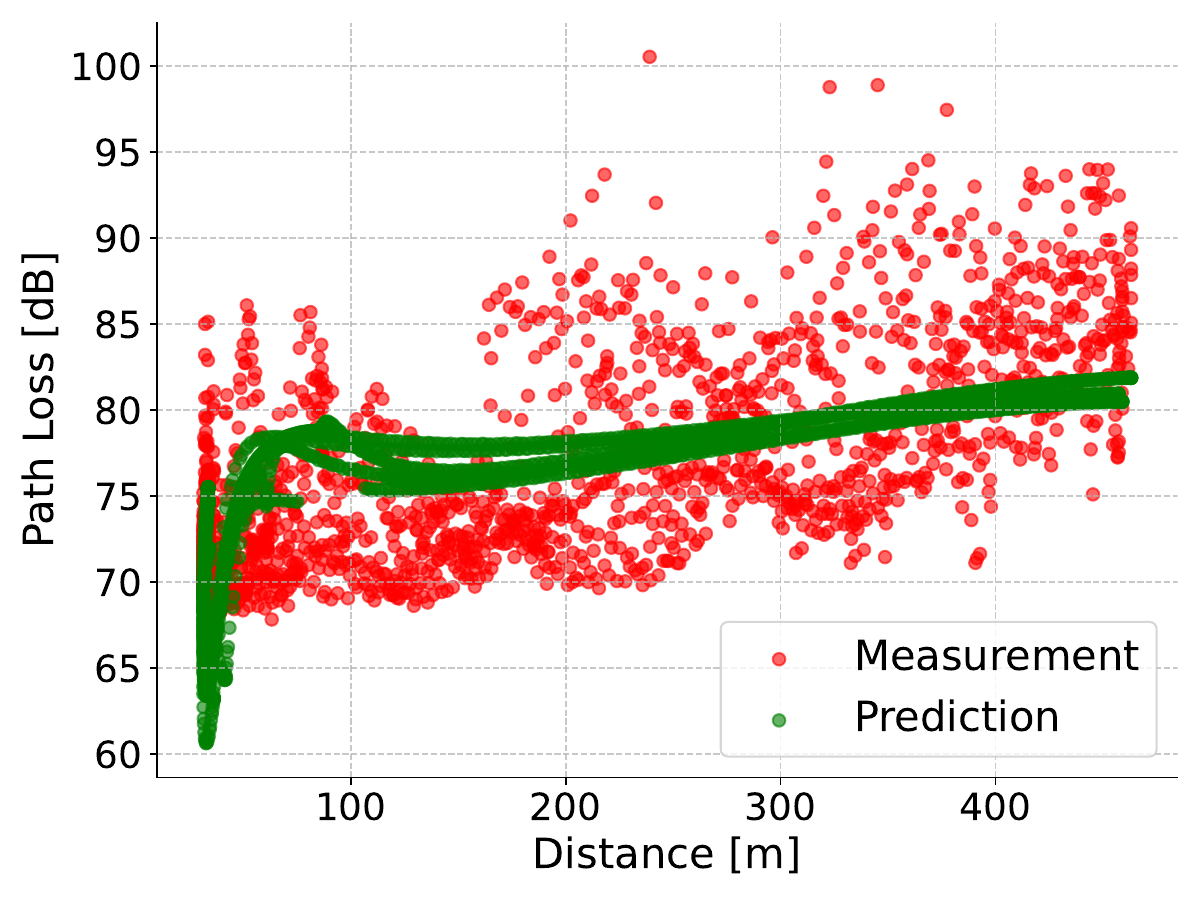}}
    \subfigure[Symbolic Base KAN: measured vs. predicted]{
        \includegraphics[width=0.22\linewidth]{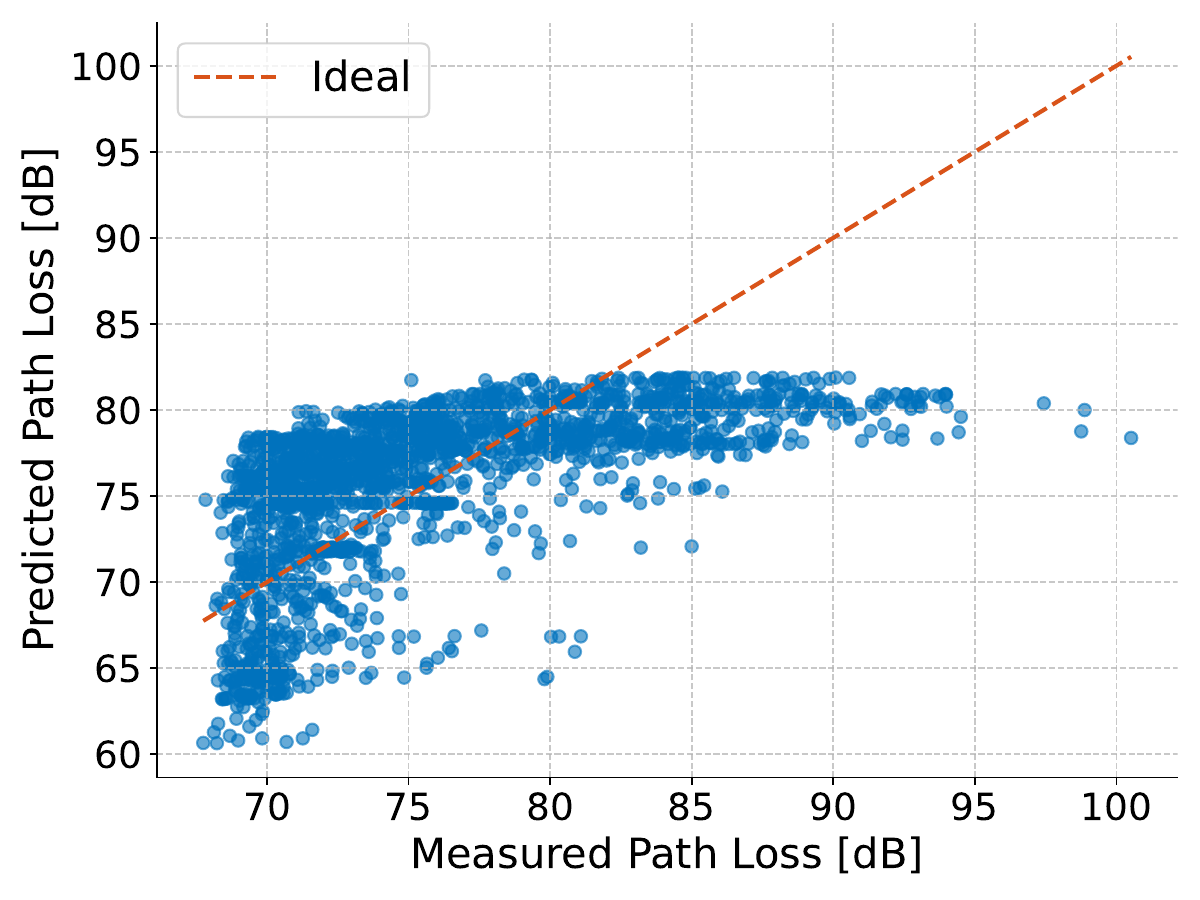}}

    \subfigure[Retrained Symbolic KAN: distance vs. path loss]{
        \includegraphics[width=0.22\linewidth]{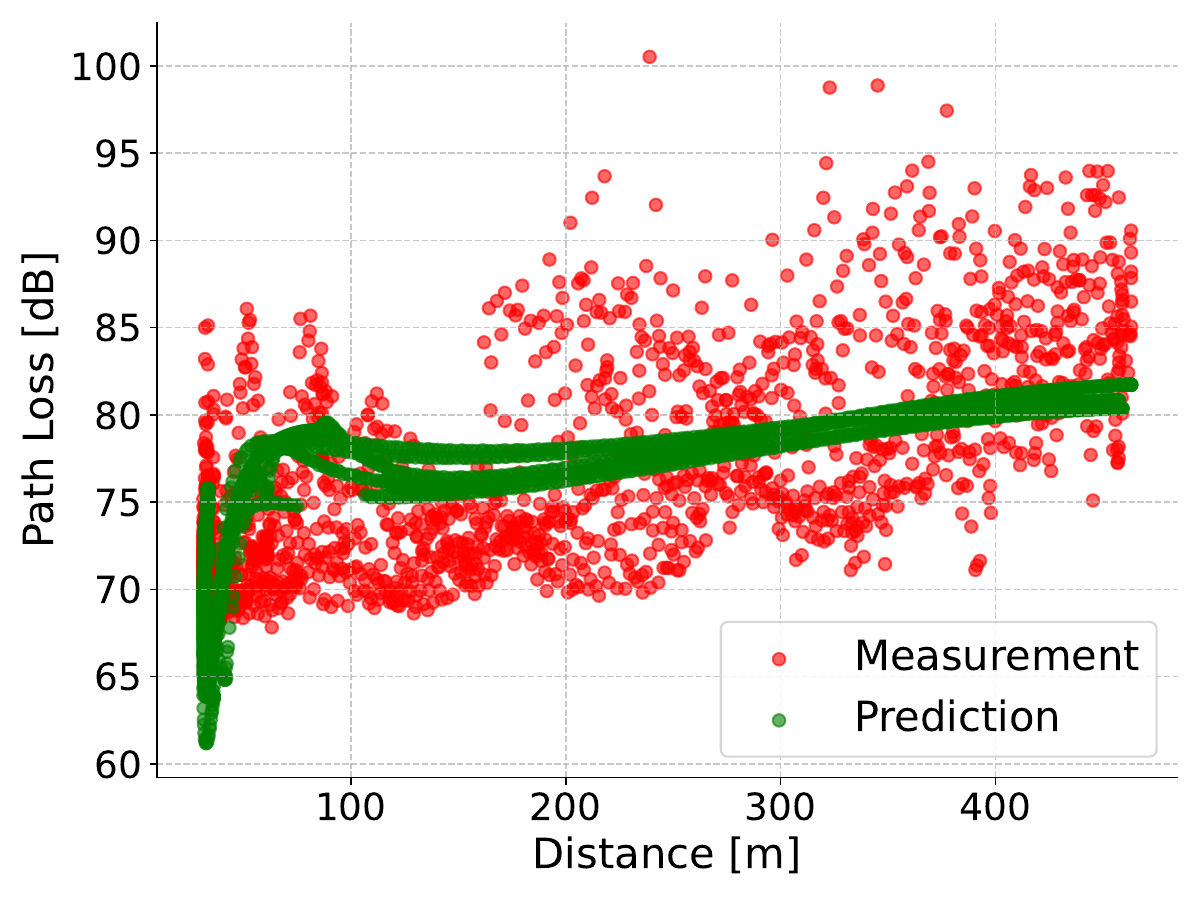}}
    \subfigure[Retrained Symbolic KAN: measured vs. predicted]{
        \includegraphics[width=0.22\linewidth]{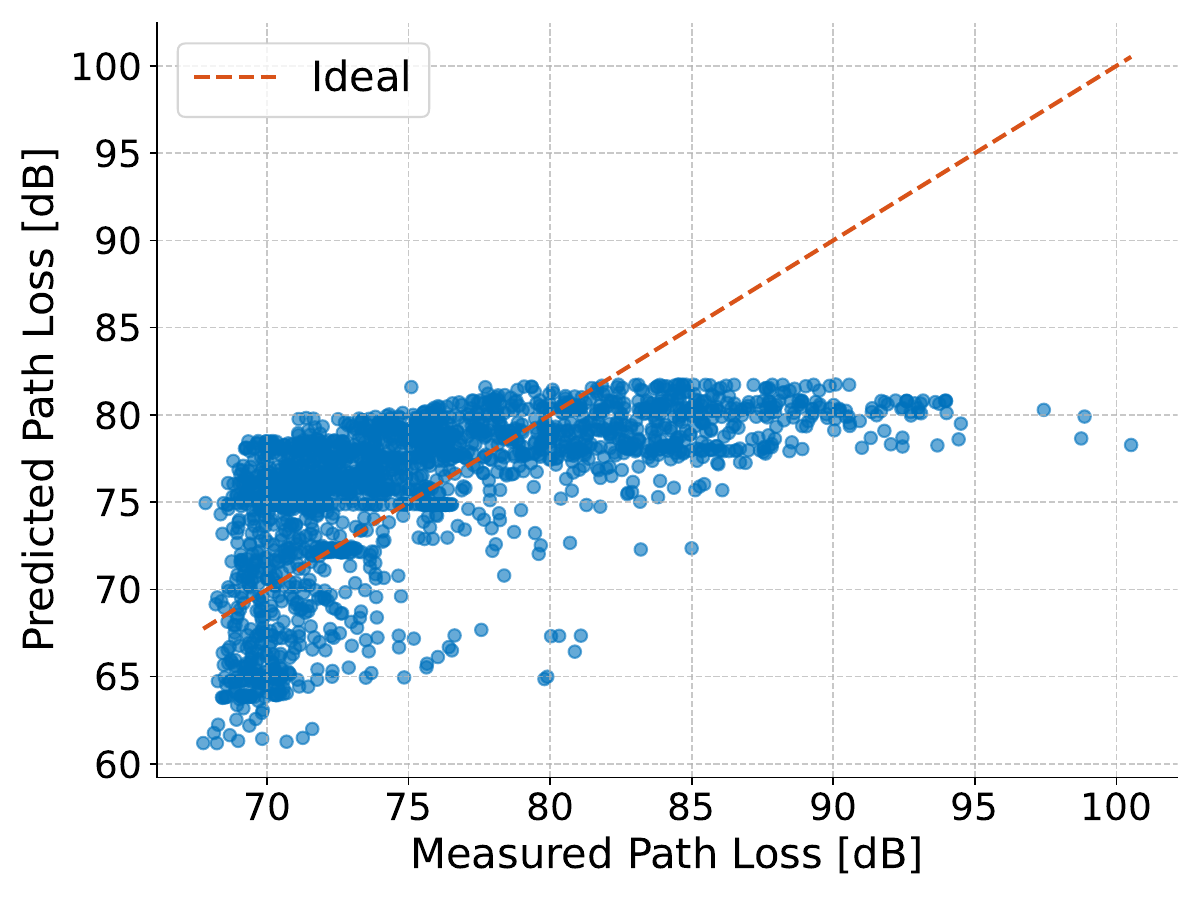}}
    \subfigure[PIKAN-FSPL: distance vs. path loss]{
        \includegraphics[width=0.22\linewidth]{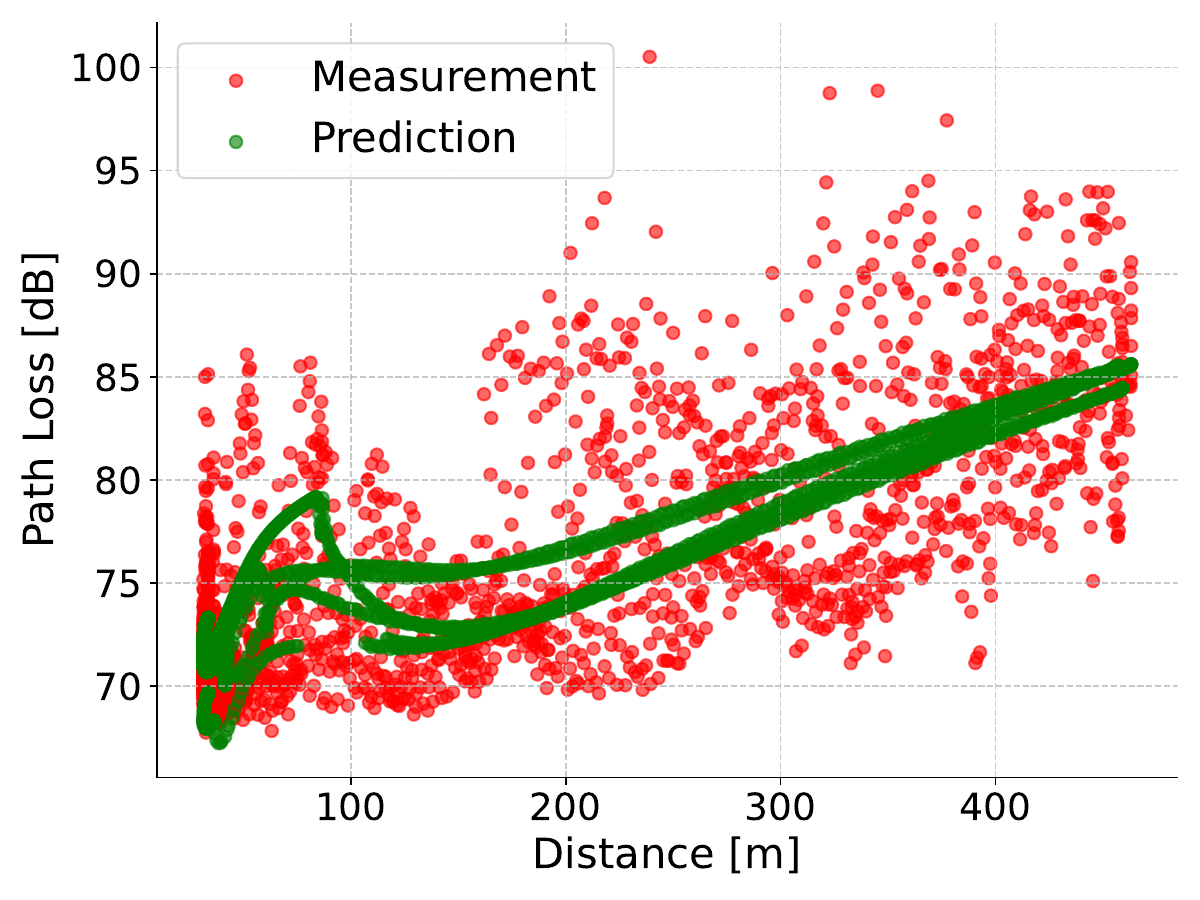}}
    \subfigure[PIKAN-FSPL: measured vs. predicted]{
        \includegraphics[width=0.22\linewidth]{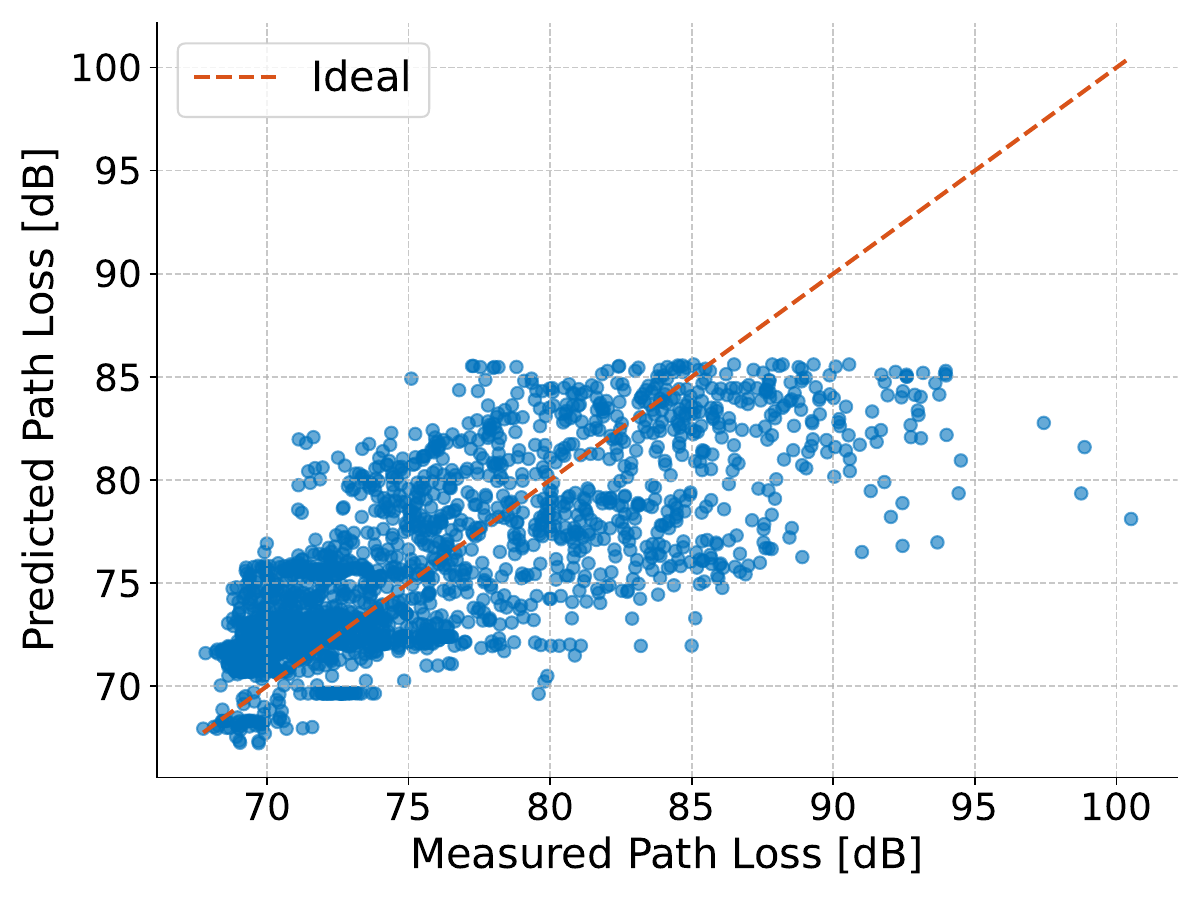}}

    \subfigure[Retrained PIKAN-FSPL: distance vs. path loss]{
        \includegraphics[width=0.22\linewidth]{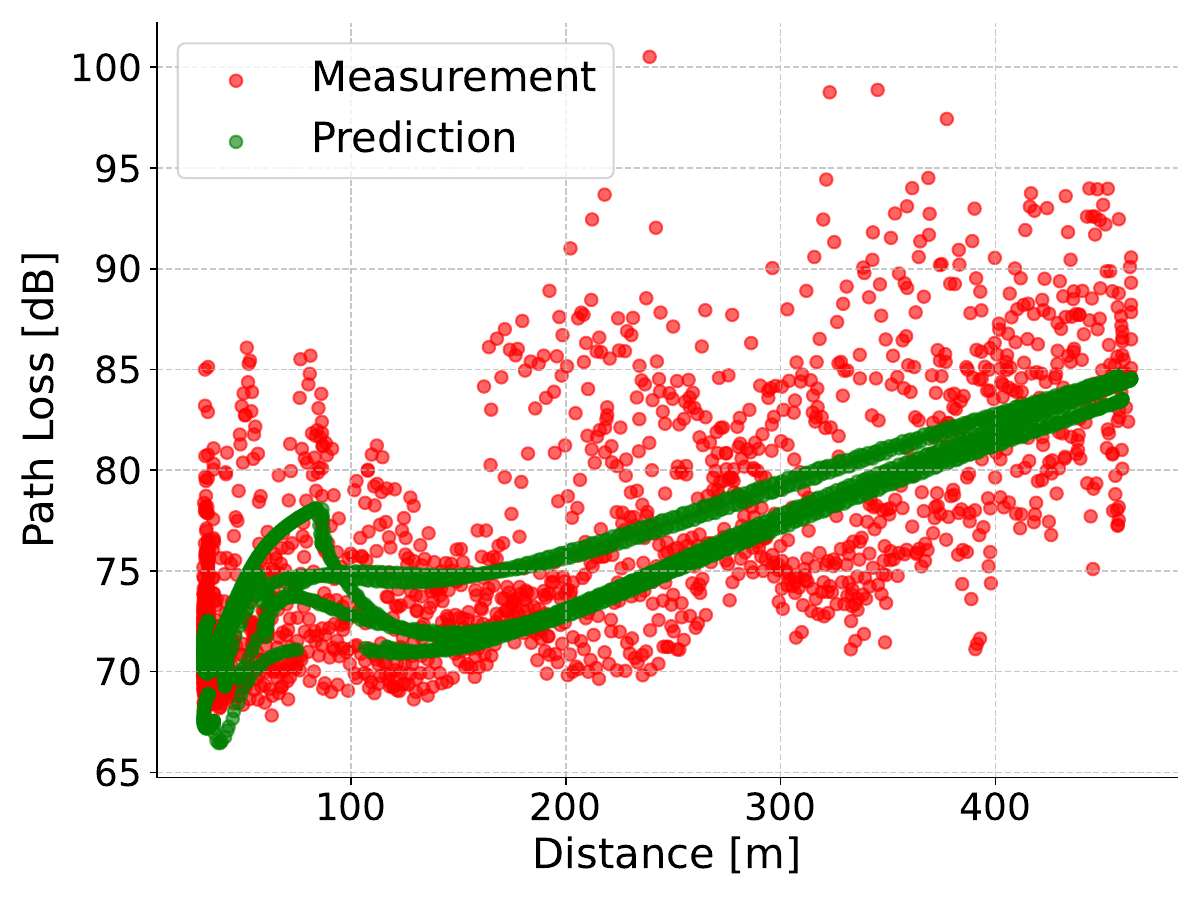}}
    \subfigure[Retrained PIKAN-FSPL: measured vs. predicted]{
        \includegraphics[width=0.22\linewidth]{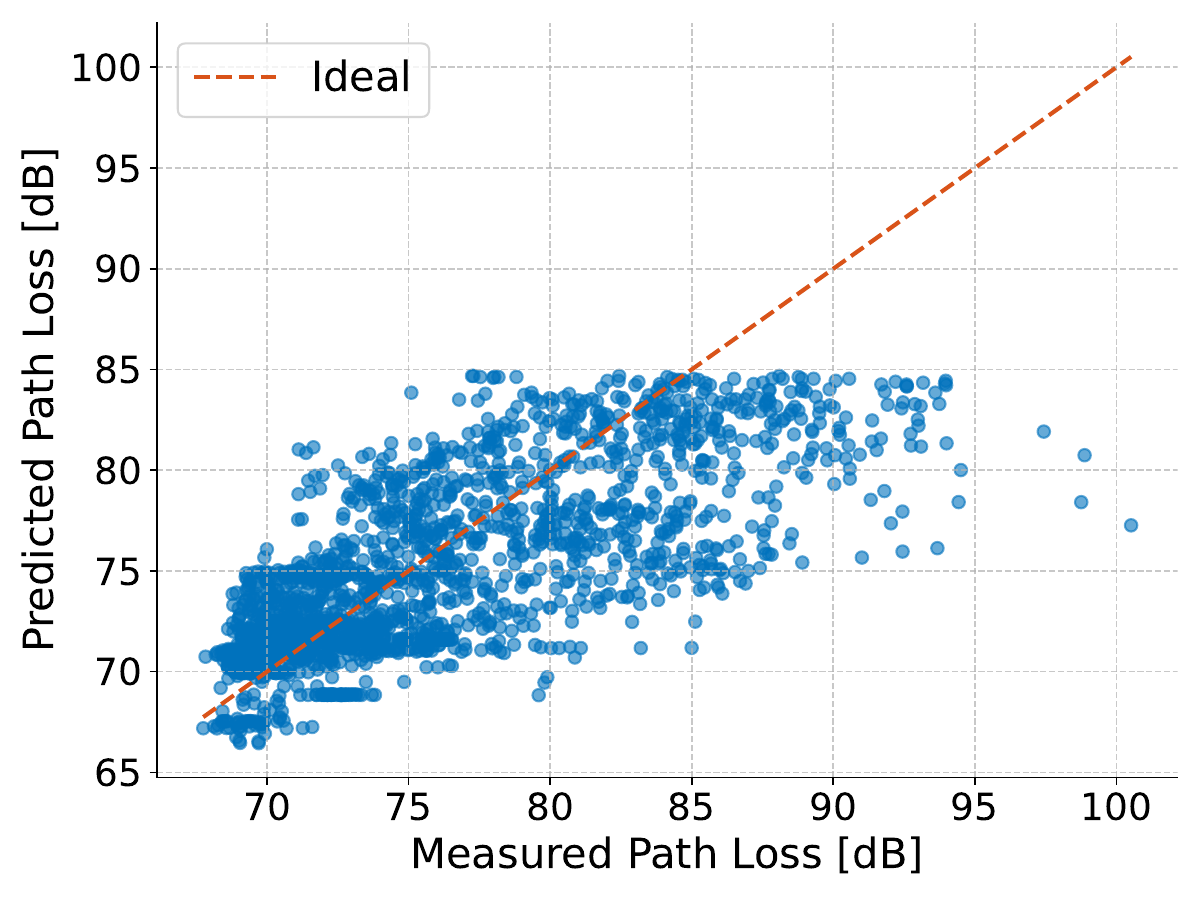}}
    \subfigure[Symbolic PIKAN-FSPL: distance vs. path loss]{
        \includegraphics[width=0.22\linewidth]{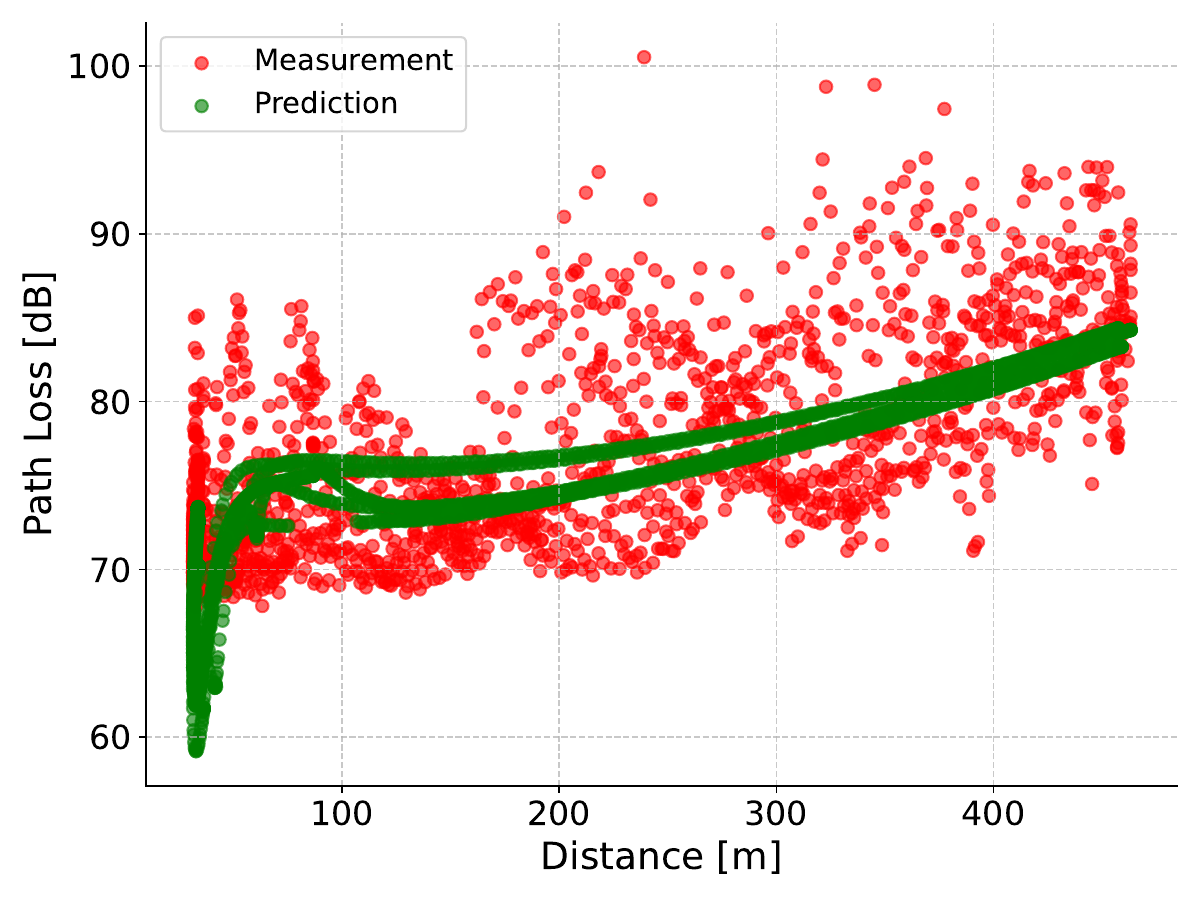}}
    \subfigure[Symbolic PIKAN-FSPL: measured vs. predicted]{
        \includegraphics[width=0.22\linewidth]{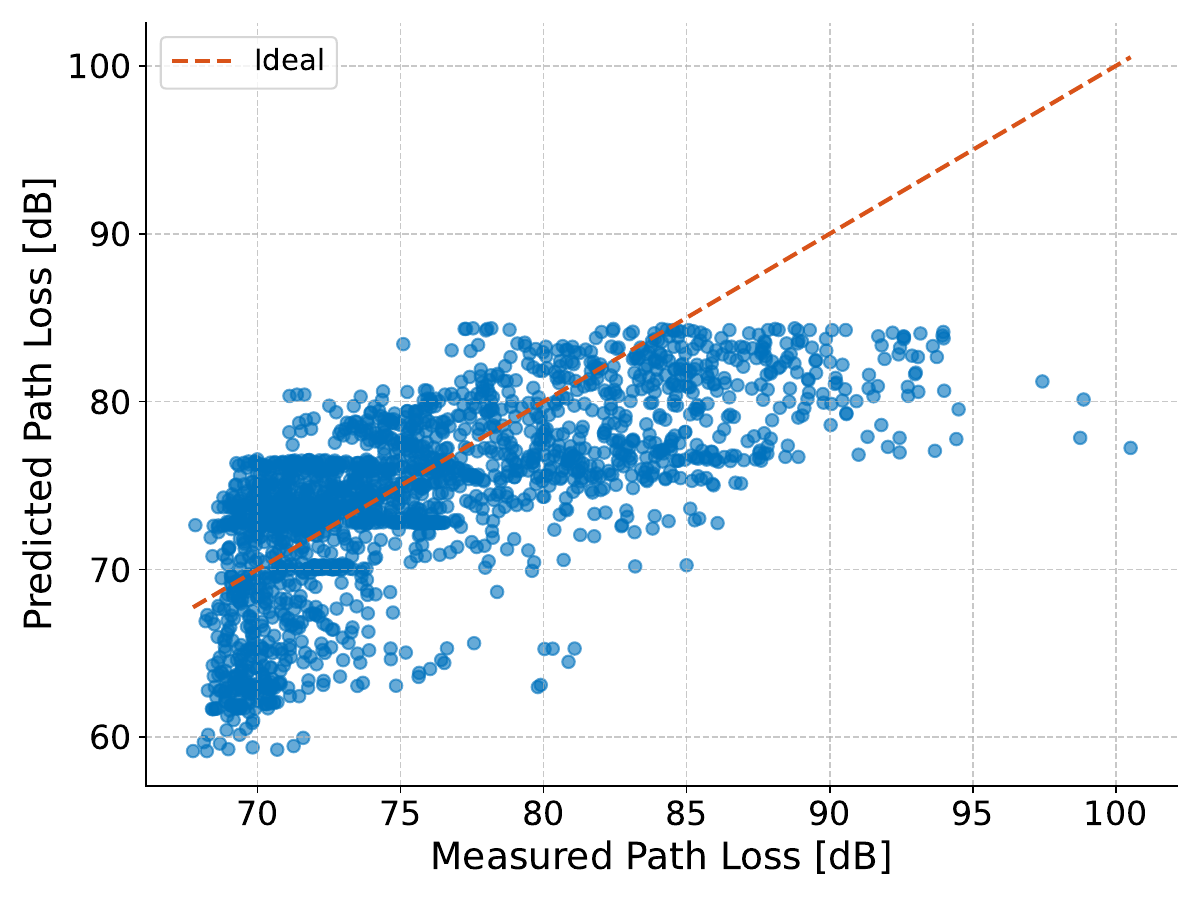}}

    \subfigure[Retrained Symbolic PIKAN-FSPL: distance vs. path loss]{
        \includegraphics[width=0.22\linewidth]{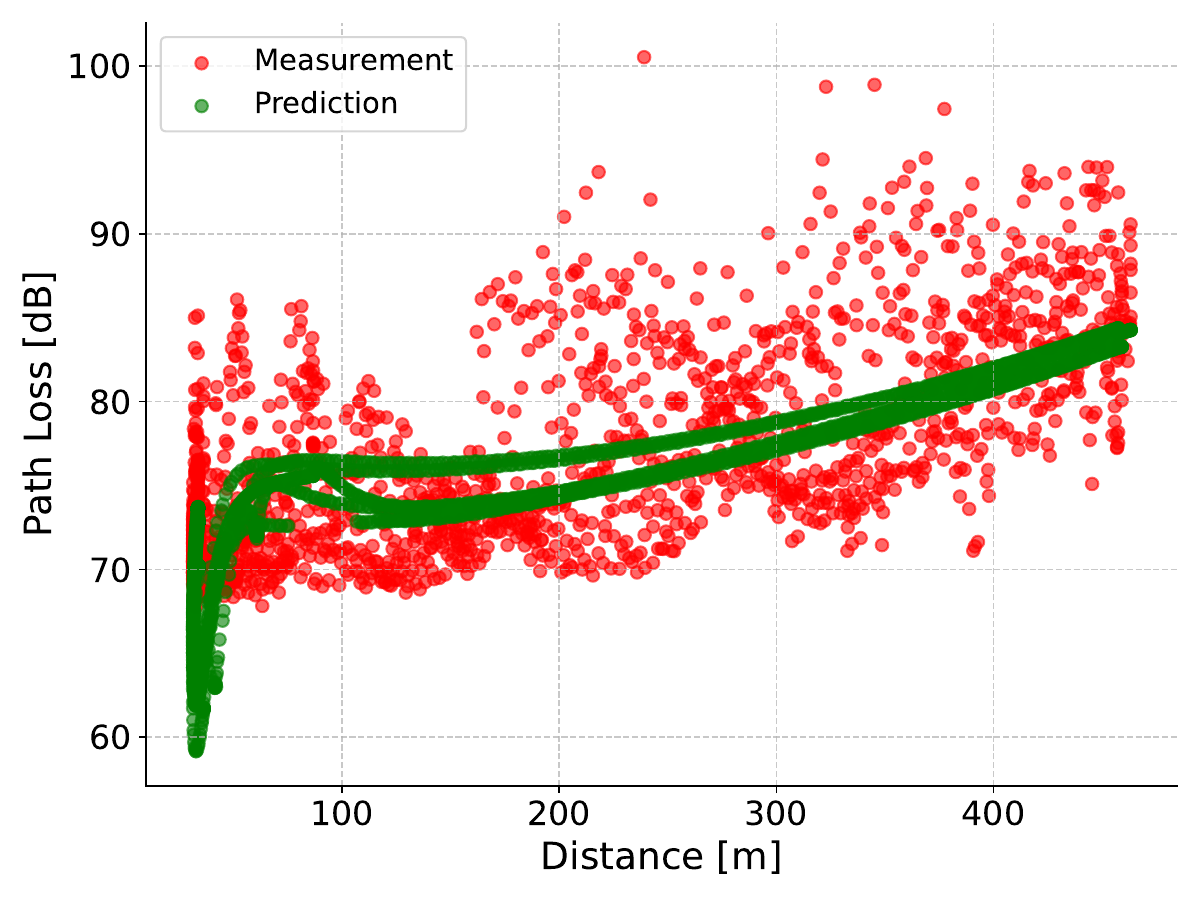}}
    \subfigure[Retrained Symbolic PIKAN-FSPL: measured vs. predicted]{
        \includegraphics[width=0.22\linewidth]{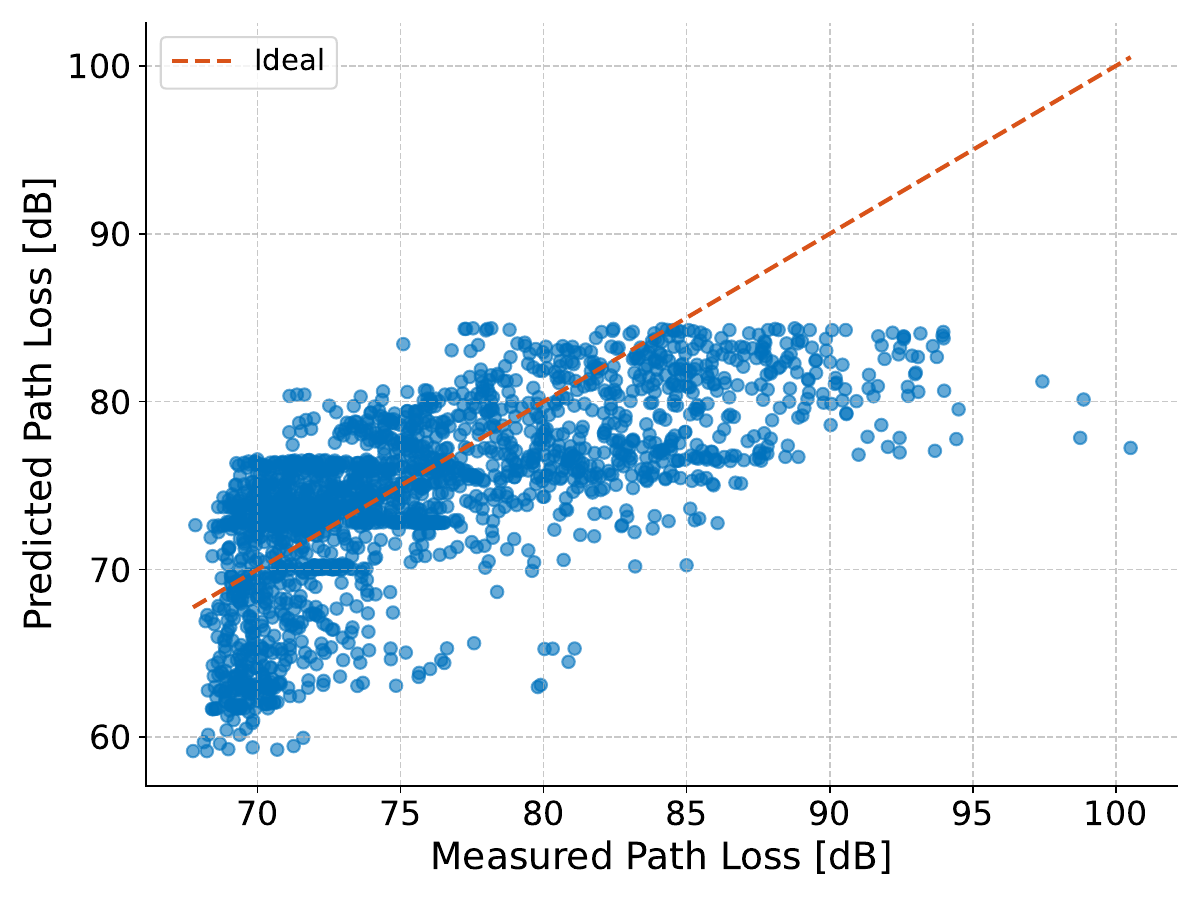}}
    \subfigure[PIKAN-2R: distance vs. path loss]{
        \includegraphics[width=0.22\linewidth]{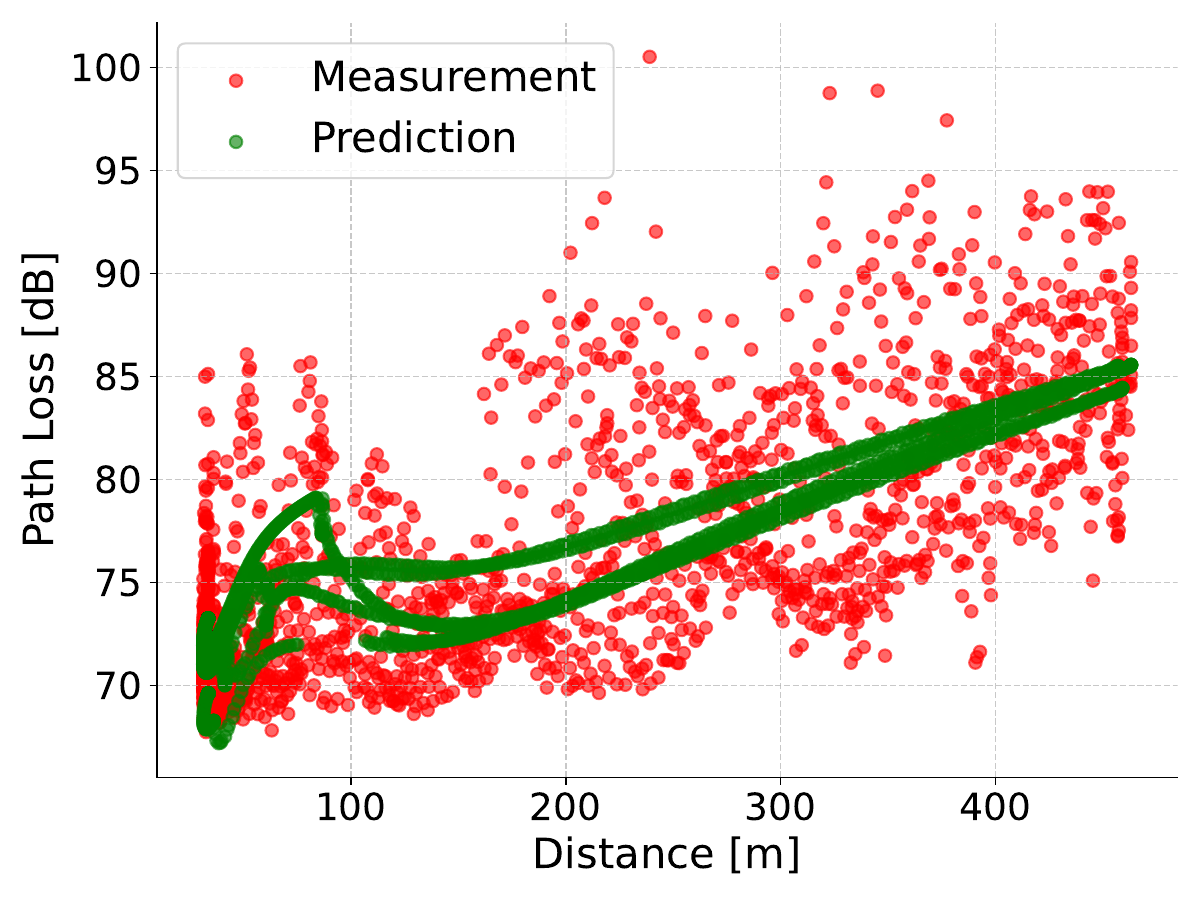}}
    \subfigure[PIKAN-2R: measured vs. predicted]{
        \includegraphics[width=0.22\linewidth]{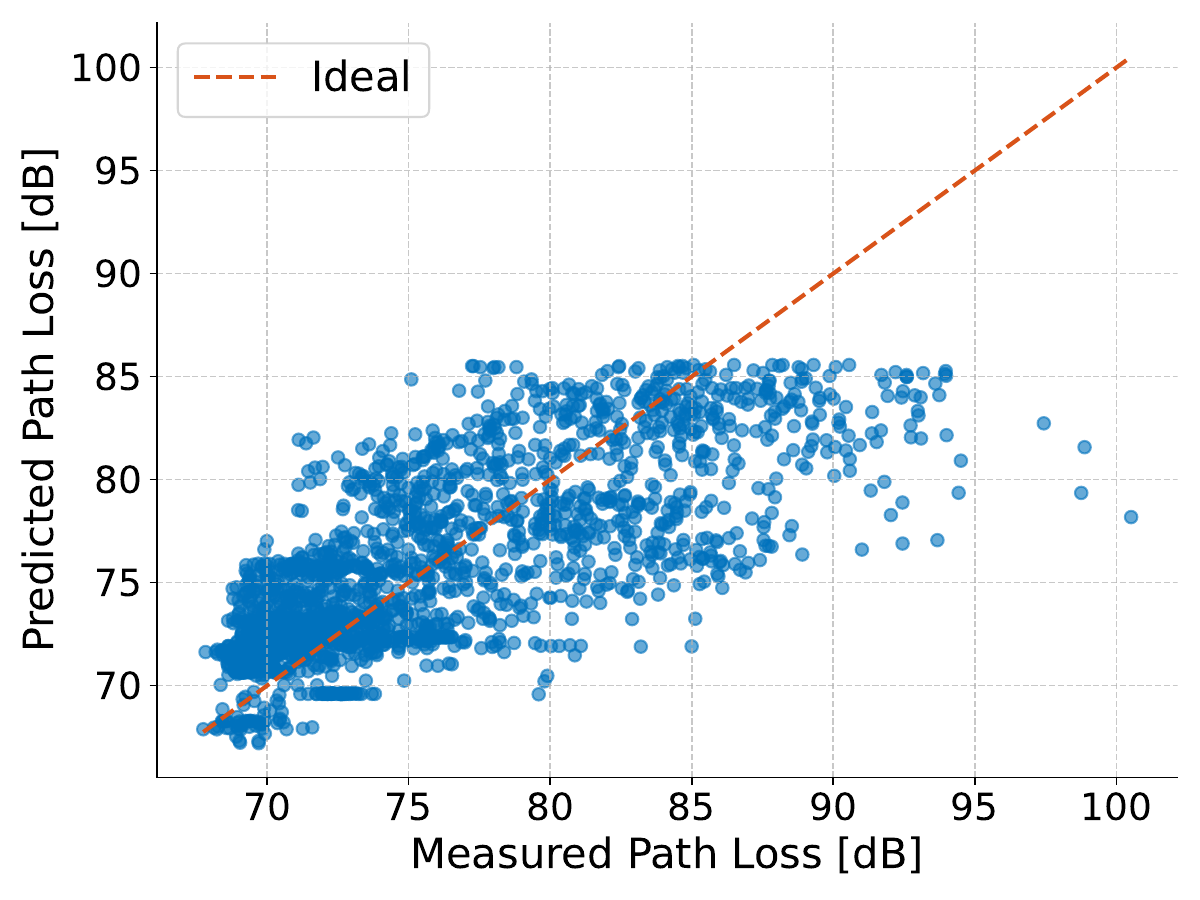}}

    \subfigure[Symbolic PIKAN-2R: distance vs. path loss]{
        \includegraphics[width=0.22\linewidth]{figs/kan_results/symbolic/scatter_distance_vs_pathloss.pdf}}
    \subfigure[Symbolic PIKAN-2R: measured vs. predicted]{
        \includegraphics[width=0.22\linewidth]{figs/kan_results/symbolic/scatter_measured_vs_predicted.pdf}}
    \subfigure[Retrained Symbolic PIKAN-2R: distance vs. path loss]{
        \includegraphics[width=0.22\linewidth]{figs/kan_results/symbolic-retrained/scatter_distance_vs_pathloss.pdf}}
    \subfigure[Retrained Symbolic PIKAN-2R: measured vs. predicted]{
        \includegraphics[width=0.22\linewidth]{figs/kan_results/symbolic-retrained/scatter_measured_vs_predicted.pdf}}

    \caption{Scatter plots comparing distance vs. path loss (left of each pair) and measured vs. predicted accuracy (right of each pair) for all KAN models listed in~\TAB{tab:kan_eval}.}
    \label{fig:kan_eval_scatters}
\end{figure*}

\end{document}